\documentclass[letterpaper]{article} % DO NOT CHANGE THIS
\usepackage[draft]{aaai2026}
\usepackage{times}  % DO NOT CHANGE THIS
\usepackage{helvet}  % DO NOT CHANGE THIS
\usepackage{courier}  % DO NOT CHANGE THIS
\usepackage[hyphens]{url}  % DO NOT CHANGE THIS
\usepackage{graphicx} % DO NOT CHANGE THIS
\usepackage[utf8]{inputenc}
\usepackage[T1]{fontenc}
\usepackage{graphicx}
\usepackage{booktabs} % for professional tables
\usepackage{microtype}

\usepackage{algorithmic}

\usepackage{xcolor}
\usepackage{algorithm}
\usepackage{amsmath}
\usepackage{amssymb}
\usepackage{mathtools}
\usepackage{amsthm}
\usepackage{booktabs}

\usepackage{xspace}
\usepackage{lipsum}
\usepackage{threeparttable}
\usepackage{tablefootnote}
\usepackage{colortbl}
\usepackage{lineno}
\usepackage{array, arydshln} % Load arydshln for dashed lines
\usepackage{float}
\usepackage{subcaption}
\definecolor{iccvblue}{rgb}{0.21,0.49,0.74}
\usepackage{natbib}  % DO NOT CHANGE THIS AND DO NOT ADD ANY OPTIONS TO IT
\usepackage{caption} % DO NOT CHANGE THIS AND DO NOT ADD ANY OPTIONS TO IT
\usepackage{algorithm}
\usepackage{algorithmic}

\usepackage{newfloat}
\usepackage{listings}
\DeclareCaptionStyle{ruled}{labelfont=normalfont,labelsep=colon,strut=off} % DO NOT CHANGE THIS
\floatstyle{ruled}
\newfloat{listing}{tb}{lst}{}
\floatname{listing}{Listing}
\definecolor{codeblue}{rgb}{0.25,0.5,0.5}
\definecolor{codekw}{rgb}{0.85, 0.18, 0.50}

\definecolor{codesign}{RGB}{0, 0, 255}
\definecolor{codefunc}{rgb}{0.0, 0.0, 0.9}

\lstdefinelanguage{PythonFuncColor}{
  language=Python,
  keywordstyle=\color{blue}\bfseries,
  commentstyle=\color{codeblue},  % for lines starting with "#"
  stringstyle=\color{orange},
  showstringspaces=false,
  basicstyle=\ttfamily\small,
  literate=
    {*}{{\color{codesign}* }}{1}
    {-}{{\color{codesign}- }}{1}
    {+}{{\color{codesign}+ }}{1}
    {dataloader}{{\color{codefunc}dataloader}}{1}
    {sample_t_r}{{\color{codefunc}sample\_t\_r}}{1}
    {randn}{{\color{codefunc}randn}}{1}
    {randn_like}{{\color{codefunc}randn\_like}}{1}
    {jvp}{{\color{codefunc}jvp}}{1}
    {stopgrad}{{\color{codefunc}stopgrad}}{1}
    {metric}{{\color{codefunc}metric}}{1}
}

\title{MolSnap: Snap-Fast Molecular Generation with Latent Variational Mean Flow}
\author{
    Md Atik Ahamed\textsuperscript{\rm 1}, Qiang Ye\textsuperscript{\rm 2}, Qiang Cheng\textsuperscript{\rm 1, \rm 3}\thanks{Corresponding author}
}
\affiliations{
    \textsuperscript{\rm 1}Department of Computer Science\\
    \textsuperscript{\rm 2}Department of Mathematics\\
    \textsuperscript{\rm 3}Institute for Biomedical Informatics\\
    University of Kentucky\\
    \{atikahamed,qye3,qiang.cheng\}@uky.edu
    
}

\begin{document}

\maketitle

\begin{abstract}
Molecular generation conditioned on textual descriptions is a fundamental task in computational chemistry and drug discovery. Existing methods often struggle to simultaneously ensure high-quality, diverse generation and fast inference. In this work, we propose a novel causality-aware framework that addresses these challenges through two key innovations. First, we introduce a Causality-Aware Transformer (CAT) that jointly encodes molecular graph tokens and text instructions while enforcing causal dependencies during generation. Second, we develop a Variational Mean Flow (VMF) framework that generalizes existing flow-based methods by modeling the latent space as a mixture of Gaussians, enhancing expressiveness beyond unimodal priors. VMF enables efficient one-step inference while maintaining strong generation quality and diversity. Extensive experiments on four standard molecular benchmarks demonstrate that our model outperforms state-of-the-art baselines, achieving higher novelty (up to 74.5\%), diversity (up to 70.3\%), and 100\% validity across all datasets. Moreover, VMF requires only one number of function evaluation (NFE) during conditional generation and up to five NFEs for unconditional generation, offering substantial computational efficiency over diffusion-based methods.
\end{abstract}

\section{Introduction}

The ability to generate novel molecular structures conditioned on textual descriptions is a key enabler for modern drug discovery and materials science~\cite{ma2021paving,southey2023introduction}. Traditional molecular design relies heavily on expert knowledge and brute-force screening of chemical libraries~\cite{pyzer2015high}, both of which are time-consuming and limited in their ability to explore the vast chemical space. Recent advances in deep generative modeling offer promising alternatives, enabling the automatic generation of diverse, valid molecules that satisfy user-defined criteria expressed in natural language~\cite{bilodeau2022generative,ilnicka2023designing}.

Despite these advances, existing approaches face several core challenges. First, most methods treat molecule generation as a generic sequence-to-sequence or graph generation task, overlooking the causal dependencies that govern molecular assembly and property formation~\cite{li2018learning,you2018graph}. In real chemical systems, the emergence of molecular properties is driven by specific causal relationships among structural motifs~\cite{hajduk2007decade}. Second, leading models often rely on diffusion-based or multi-step processes that require many inference iterations, making them computationally expensive~\cite{ho2020denoising,zhu20243m}. Third, flow-based models typically assume unimodal Gaussian priors~\cite{madhawa2019graphnvp,zang2020moflow}, which may fail to capture the inherently multimodal nature of molecular distributions, thereby limiting the diversity and quality of generated samples.

To overcome these limitations, we introduce a novel framework that combines causality-aware modeling with variational flow matching for efficient and expressive molecular generation. Motivated by the observation that structural features, such as functional groups, causally influence key molecular properties like reactivity, solubility, and bioactivity~\cite{mandal2009rational,bickerton2012quantifying}, our framework incorporates causal reasoning for both model architectures and generation dynamics.

\begin{figure}[t]
    \centering
    \includegraphics[width=0.7\linewidth]{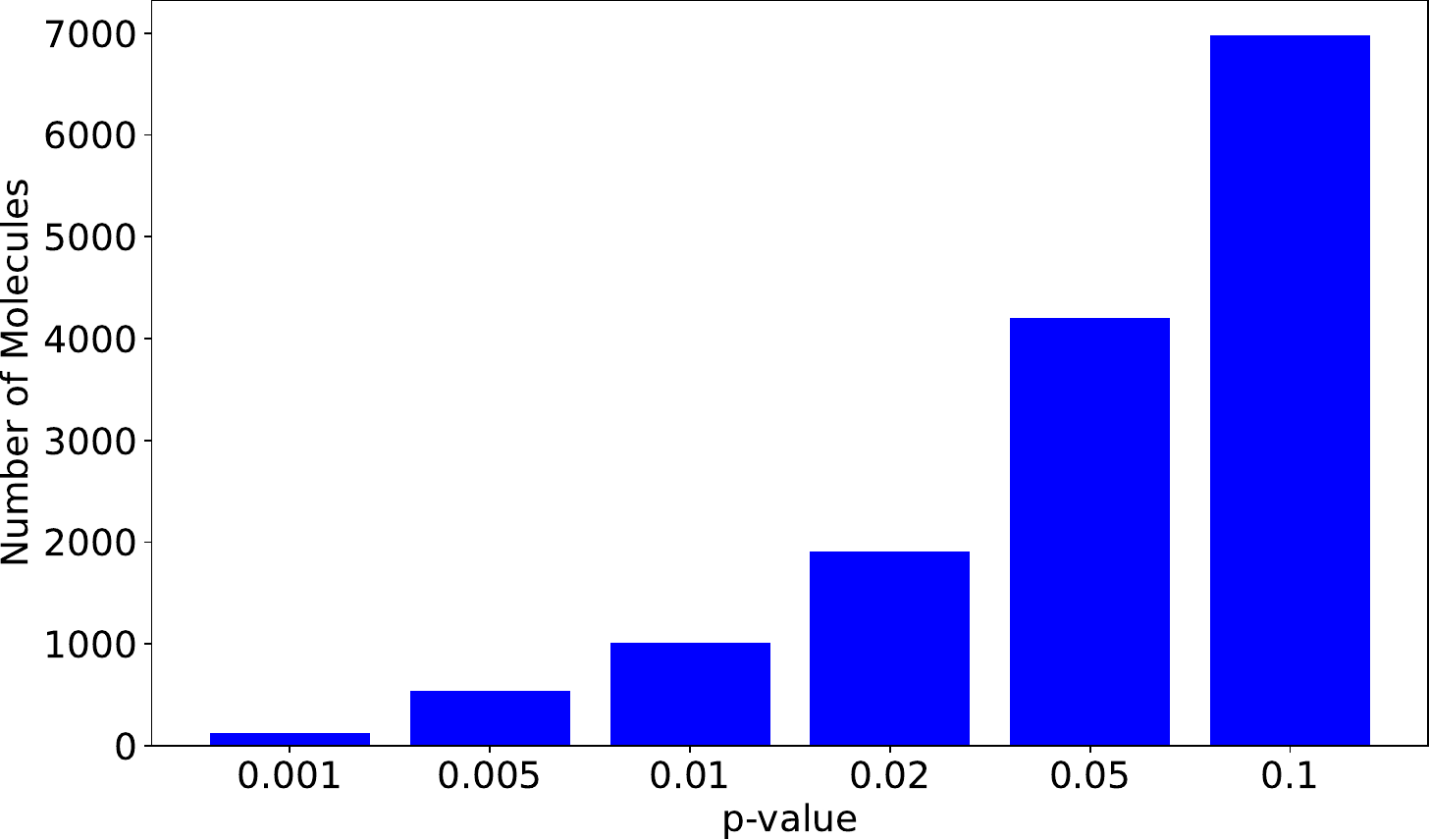}
    \caption{Causality analysis via Granger causality test verifying motivation for utilizing causality in our architecture.}
    \label{fig:causality}
\end{figure}

Our approach builds on a shared latent space where molecular graphs and textual instructions are encoded to be causality aware through modality-specific converters, enabling seamless integration of structural, semantic information and causal relationships between molecular components. To validate this design, we conducted latent causality analysis on ChEBI-20 using Granger tests consistent with our causal formulation (Figure~\ref{fig:causal_dependency}). As illustrated in Figure~\ref{fig:causality}, numerous molecules exhibit strong causal links ($p<0.001$), and nearly half show at least one link at $p<0.1$. These findings confirm that the latent space captures structured molecular causality effectively leveraged by our attention mask. Building on this insight, we introduce the Causality-Aware Transformer (CAT), which enforces directional dependencies through masked attention to ensure causally coherent generation of molecular substructures.

In addition, we propose a variational mean flow (VMF) framework that extends traditional mean flow models~\cite{geng2025mean}. By modeling the latent space as a mixture of Gaussians, VMF captures the multimodal nature of molecule distributions and enables efficient one-step inference. This variational formulation promotes generation diversity while maintaining structural validity and semantic alignment with input instructions.

\vspace{0.5em}
In summary, our key contributions include:
\begin{itemize}
    \item We introduce a causality-aware Transformer (CAT) that explicitly models dependencies among molecular graph tokens and text instruction tokens.
    \item We propose a variational mean flow (VMF) framework that generalizes standard flow matching and rectified flow models for molecule generation.
    \item VMF models the latent space as a mixture of Gaussians, improving diversity and capturing multimodal distributions more effectively than unimodal priors.
    \item Our model supports fast generation with only one or a few number of function evaluations (NFEs), significantly reducing inference cost compared to diffusion-based methods.
\end{itemize}

We validate our method on four standard molecular benchmarks, demonstrating that MolSnap achieves superior performance in terms of novelty, diversity, and validity, while also offering substantial gains in inference efficiency over state-of-the-art baselines.

\section{Related Work}
\noindent\textbf{Molecular Representation and Generation.}
Molecular generation has progressed from SMILES-based sequence models~\cite{weininger1988smiles,bjerrum2017molecular,gomez2018automatic,kusner2017grammar} to graph-based approaches that better capture molecular structures~\cite{li2018learning,you2018graph,liu2018constrained,jin2018junction,jin2020hierarchical}. While string-based methods face challenges with validity, graph-based models improve chemical correctness and interpretability but may lack diversity and scalability.
Recent multimodal approaches leverage textual descriptions for molecule generation~\cite{edwards2021text2mol,edwards2022translation,schwaller2019molecular,born2023regression,fang2023mol}, showing promise but often requiring slow multi-step inference and lacking explicit causal modeling.\\

\noindent\textbf{Flow-Based Generative Models.}
Flow-based models enable exact likelihood estimation and stable training~\cite{madhawa2019graphnvp}, with applications in molecular generation such as GraphNVP~\cite{madhawa2019graphnvp}, MoFlow~\cite{zang2020moflow}, and GraphDF~\cite{luo2021graphdf}. Recent advances in flow matching and mean flow modeling~\cite{geng2025mean} simplify training and improve efficiency. However, most methods rely on unimodal Gaussian priors, limiting expressiveness for multimodal molecular distributions. Variational extensions like VRFM~\cite{guovariational} improve diversity but are only explored in vision domains. Our work brings these advances to molecular generation while introducing causality-aware components.\\

\noindent\textbf{Causality in Generative Modeling.}  
Causal generative models~\cite{deng2024causal} aim to uncover underlying mechanisms beyond mere correlations, which is essential for capturing functional group interactions and structure–property relationships in molecules~\cite{hajduk2007decade,mandal2009rational}. However, prior work in causal representation learning has primarily focused on image domains and does not address the sequential nature of molecular assembly. Our causality-aware Transformer (CAT) fills this gap by enforcing autoregressive dependencies through masked attention, thereby enabling both structured generation and interpretability.\\

\noindent\textbf{Conditional Molecular Generation.}
Conditional generation, whether guided by properties or text, is essential for controllable drug design~\cite{westermayr2023high,schneuing2024structure}. Property-conditioned VAEs~\cite{gomez2018automatic}, Transformer-based models~\cite{christofidellis2023unifying,liu2023molca}, and diffusion approaches~\cite{zhu20243m,weiss2023guided} have made progress, but suffer from inference inefficiency. Models like Mol-Instructions~\cite{fang2023mol} enable natural language control but do not capture causal structure-property relationships.
Our framework uniquely combines causal modeling with variational mean flow, enabling fast, high-quality conditional generation with interpretable dynamics.

\section{Methodology}
In this section, we outline our methodology. Given a training set of molecule–text pairs, the goal is to learn a conditional generative model that produces valid, diverse molecules aligned with the text.

\begin{figure}[hbt]
    \centering
    \includegraphics[width=0.96\linewidth]{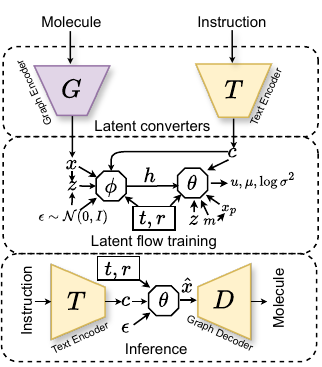}
    \caption{Overview of our method. Converters $G$ and $T$ map molecules and instructions into latents. Latent flow training learns generative dynamics in this space, while the inference module produces molecules conditioned on instructions.}
    \label{fig:molsnap}
\end{figure}
Each molecule is represented as a graph $g = (V, E)$, where $V = \{v_1, v_2, \dots, v_{|V|}\}$ is the set of atoms/nodes and $E$ is the set of chemical bonds/edges. The atom information is stored in a matrix $N \in \mathbb{R}^{|V| \times D}$, where each row encodes properties such as atom type and chirality. The bonding structure is described by an adjacency tensor $A \in \mathbb{R}^{|V| \times |V| \times b}$, where each slice specifies the bond types. Our overall scheme is demonstrated in Figure~\ref{fig:molsnap}.

\subsection{Latent Converters}
We refer to $G$ and $T$ as latent converters since they transform high-dimensional, unstructured graph and text data into compact latent representations. Specifically, $G$ encodes a molecule graph $g$ into a latent representation $x$, while $T$ maps a conditional instruction $i$ into a latent representation $c$. To ensure a shared latent space, we first apply contrastive alignment training, which enforces similarity between paired $(g, i)$ representations and dissimilarity among unrelated pairs. Additionally, we train a graph encoder–decoder framework, where the decoder $D$ reconstructs molecules from latent vectors. During inference, $D$ is used to recover the final molecular graph from the predicted latent representation $\hat{x}$.

\subsection{Latent Flow Training}
Given the latent representations $x \gets G(g)$ and $c \gets T(i)$, we jointly train the networks $\theta$ and $\phi$. The overall training process is outlined in Algorithm~\ref{alg:training} for a single example. Our approach differs fundamentally from prior works: rather than relying on conventional diffusion models~\cite{ho2020denoising}, we perform training directly in the latent space utilizing Mean Flow~\cite{geng2025mean},  VRFM~\cite{guovariational}, and CAT~\cite{deng2024causal}. 

While the original Mean Flow was introduced for image data, we are the first to adapt and extend it to molecule–text modalities. This extension enables us for efficient 1-NFE molecular generation. While Mean Flow provides an efficient framework, we noticed that its latent space is modeled with unimodal gausian distribution. However, this does not address the multi-modality issue identified in Variational Rectified Flow Matching (VRFM)~\cite{guovariational}. 

To capture multi-modality, we study the use of a mixture model over
velocities at each data-domain-time-domain location. Specifically, inspired by VRFM~\cite{guovariational}, we utilize a KL-Divergence term (Specific formulation is described later in this Section) as a regularization in our objective function together with Mean Flow based cost. Moreover, we go beyond a naive adaptation by integrating Mean Flow with Variational Rectified Flow Matching (VRFM)~\cite{guovariational}. VRFM was initially designed for image modalities, and our framework is the first to combine VRFM with Mean Flow for molecular generation, demonstrating a novel and effective integration.

To achieve this, we employ two core networks: a causality-aware Transformer (CAT), denoted as $\theta$ in Algorithm~\ref{alg:training} and Algorithm~\ref{alg:inference}, and a variational encoder $\phi$. The encoder $\phi$ takes as input the condition $c$, noise $\epsilon$, clean latent $x$, intermediate latent $z$, and timesteps $t$ and $r$.

Its functionality is summarized by the following equations:

\begin{align}
    \mu, \log \sigma^2 &= \phi(c, \epsilon, x, z, r, t), \\
    \sigma &= e^{\frac{1}{2} \log \sigma^2}, \\
    h &= \mu + \sigma \odot \epsilon, \quad \text{where } \epsilon \sim \mathcal{N}(0, I).
    \label{eq:phi_working}
\end{align}
After obtaining the latent representation $h$ from $\phi$, we combine it with the partial clean input $x_p$, causal mask $m$, and with other inputs $c, z, t, r$. The roles of causal masking and the partial input $x_p$ are discussed later in this subsection. The network $\theta$ processes these inputs to produce $u$, which represents the predicted velocity.

To compute both the instantaneous velocity $u$ and its derivative with respect to time, we employ the Jacobian-vector product (JVP), which efficiently estimates directional derivatives. In our setting, JVP operates on the composite network $(\theta, \phi)$ with input $(z, r, t)$ and tangent vector $(v, 0, 1)$, as illustrated in Algorithm~\ref{alg:training}. 
where $v$ is the target vector. Here, $u $ represents the predicted velocity, while $\dot{u} = du/dt$ denotes its derivative with respect to the time variable $t$. With the help of $v,t,r,\dot{u}$, we compute the target velocity $u_t$ as demonstrated in Algorithm~\ref{alg:training}.

In addition to the $L_2$ loss we also incorporate KL-Divergence loss $\mathcal{L}_{KL}$ and Dispersive loss~\cite{wang2025diffuse} $\mathcal{L}_{disp}$ and combine them with weight $\alpha$ and $\beta$ respectively to formulate the final loss $\mathcal{L}$. 

To regularize the latent space $h$ produced by $\phi$, we introduce a Kullback–Leibler (KL) divergence that aligns the latent distribution with a standard Gaussian prior $\mathcal{N}(0, I)$. Specifically, given $\mu$ and $\log \sigma^2$ from $\phi$, the KL loss is computed as:
\begin{equation}
    \mathcal{L}_{KL} = \frac{1}{2} \sum_j \left( e^{\log \sigma_j^2} + \mu_j^{2} - 1 - \log \sigma_j^{2} \right).
\end{equation}
where the summation runs over latent dimensions $j$. This term encourages the latent variables to follow a unit Gaussian, improving stability and enabling a variational formulation.

To promote diversity in the latent space, we incorporate a dispersive loss $\mathcal{L}_{\text{disp}}$, defined as:
\begin{equation}
    \mathcal{L}_{\text{disp}} = \log \left( \frac{1}{B^2} \sum_{b_1,b_2} \exp\!\left( - \frac{\| z_{b_1} - z_{b_2} \|_2^2}{\tau} \right) \right),
\end{equation}
where $z_{b_1}$ and $z_{b_2}$ are latent representations in a batch of size $B$, and $\tau > 0$ is a temperature parameter controlling the sharpness of the distance penalty.

\begin{algorithm}[H]
\caption{Training}
\label{alg:training}
\begin{algorithmic}[1]
    \STATE $\epsilon \sim \mathcal{N}(0, I)$
    \STATE Sample $t, r$ \hfill {\color{gray}\texttt{\% Uniform or log-normal}}
    \STATE $z \gets (1 - t) \cdot x + t \cdot \epsilon$
    \STATE $v \gets \epsilon - x$
    \STATE $(u, \dot{u}, \mu, \log \sigma^2) \gets \texttt{jvp}((\theta, \phi),(z,r,t), (v, 0, 1))$ 
    
           \hfill {\color{gray}\texttt{\% $h,\mu, \log \sigma^2\gets\phi$ $\gets$ $(c,\epsilon,x,z,t,r)$}} \\
            \hfill {\color{gray}\texttt{\%$u\gets\theta\gets$  $(c,h,x_p,z,m,t,r)$}}
    
    \STATE $u_t \gets v - (t - r) \cdot \dot{u}$ \hfill {\color{gray}\texttt{\% target}}
    \STATE $\mathcal{L} \gets \| u - \texttt{stopgrad}(u_t) \|^2 + \alpha \mathcal{L}_{KL} + \beta \mathcal{L}_{disp}$
\end{algorithmic}
\end{algorithm}

\paragraph{Causality integration.}
Motivated by the causality analysis in Figure~\ref{fig:causality}, we integrate a causal attention mechanism into $\theta$, making it a \textit{causality-aware Transformer} (CAT). The inputs to $\theta$ are concatenated along the token dimension, enabling dynamic context modeling during training and decoder-like behavior during inference, where only $\epsilon$ and $c$ are required.

\begin{figure}[H]
    \centering
    \includegraphics[width=0.6\linewidth]{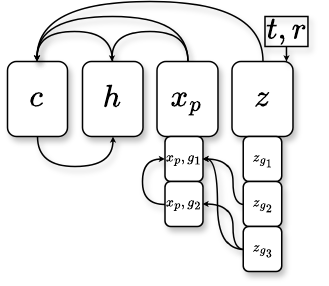}
    \caption{Causality-aware grouping of latent tokens.}
    \label{fig:causal_dependency}
\end{figure}

As shown in Figure~\ref{fig:causal_dependency}, causal dependencies are enforced via an attention mask $m$. The clean latent $x$ and noisy latent $z$ are partitioned into groups (e.g., $g_1, g_2, g_3$), where $x_p$ refers to the partial clean tokens (e.g., the first two groups). All tokens of $x_p$ and $z$ attend to the condition $c$ and latent encoding $h$, but the current group $x_{p,g_i}$ only attends up to its preceding groups $x_{p,g_{i-1}}$. For $z$, temporal information $(t, r)$ is incorporated, and causal masking ensures that each group depends only on earlier groups of $x_p$. This setup resembles autoregressive modeling in language tasks, where future tokens are masked to ensure strictly causal learning.

During inference, the clean latent tokens $x$ and $x_p$ are unavailable, as they are derived from the original data. Therefore, $\theta$ is designed to work solely with $\epsilon$ and $c$ at test time. By appending all inputs along the sequence dimension during training, the model learns to rely on the dynamic combination of noise $\epsilon$ and textual condition $c$ for reconstruction. Consequently, $\theta$ naturally functions in a decoder-only manner during inference, where the learned causal dependencies allow it to sequentially generate the latent representation without requiring $x$ or $x_p$.

\subsection{Inference}
During inference, we start with a Gaussian noise sample 
$\epsilon \sim \mathcal{N}(0, I)$ and iteratively update the latent variable by predicting 
the velocity $u = \theta(c, \epsilon, t, r)$ using the network $\theta$, then adjusting 
the latent as $\hat{x} \leftarrow \hat{x} - h \cdot u$ at each step, where $h = t - r$. For conditional generation, we leverage a one-step (1-NFE) inference strategy (Algorithm~\ref{alg:inference}), which makes the process significantly faster compared to other transformer variants that typically requires many or even 1000 NFEs.

To control the balance between conditional and unconditional signals, we employ classifier-free guidance (CFG). Specifically, we compute both a conditional output $u_{\text{cond}} = \theta(c, \epsilon, t, r)$ and an unconditional output $u_{\text{uncond}} = \theta(c_{null}, \epsilon, t, r)$, where $c_{null}$ represents a null condition. The final guided velocity is then represented by equation~\ref{eq:cfg}:
\begin{equation}
    u_{\text{final}} = w \, u_{\text{cond}} + (1 - w) \, u_{\text{uncond}}
    \label{eq:cfg}
\end{equation}
where $w$ is the guidance scale. For unconditional generation, $c$ is replaced with $c_{null}$, and $\theta$ operates solely on the noise input $\epsilon$ to produce valid molecular structures. Since our latent space is regularized by a KL divergence term to follow a standard Gaussian prior, we require only a single $\epsilon$ sample, without needing multiple independent noise components.

\begin{algorithm}[H]
\caption{Inference (1-NFE)}
\label{alg:inference}
\begin{algorithmic}[1]
    \STATE $\epsilon \sim \mathcal{N}(0, I)$
    \STATE $\hat{x}\gets\epsilon-\theta(c,\epsilon,r=0,t=1)$
    
\end{algorithmic}
\end{algorithm}

\subsection{Flow-based Models as Special Cases}
Our Variational Mean Flow (VMF) framework generalizes several existing flow-based generative models, including Mean Flow (MF) \citep{geng2025mean}, Flow Matching (FM) \citep{lipman2022flow}, and Rectified Flow Matching (RFM) \citep{liu2022rectified}. Specifically, VMF extends MF by incorporating a variational inference approach with a mixture-of-Gaussians prior, capturing multimodal latent distributions more effectively than the standard unimodal Gaussian used in MF. Flow Matching becomes a special case within VMF when adopting a unimodal prior and bypassing variational inference, simplifying the model to direct velocity matching \citep{lipman2022flow}. Similarly, RFM can be viewed as a variant of VMF where rectification adjustments are implicitly modeled by the variational posterior distribution, allowing corrections and improved stability during inference \citep{liu2022rectified}. By integrating these frameworks, VMF significantly enhances expressiveness, resulting in improved molecular generation quality, greater diversity, and more efficient inference.

\section{Experiments}
\begin{table*}
\centering

\begin{threeparttable}
\setlength{\tabcolsep}{3mm}
\begin{tabular}{l | c c c c c | c c c c c}
    \toprule
    & \multicolumn{5}{c|}{\textbf{ChEBI-20}} & \multicolumn{5}{c}{\textbf{PubChem}} \\
    Methods & Sim.& Nov.& Div.& Val.& Ove.& Sim.& Nov.& Div.& Val.& Ove.\\
    \midrule
    MolT5-small & 73.32 & 31.43 & 17.22 & 78.27 & 50.06 & 68.36 & 20.63 & 9.32 & 78.86 & 44.29 \\
    MolT5-base  & 80.75 & 32.83 & 17.66 & 84.63 & 53.97 & 73.85 & 21.86 & 9.89 & 79.88 & 46.37\\
    MolT5-large & 96.88 & 12.92 & 11.20 & 98.06 & 54.77 & 91.57 & 20.85 & 9.84 & 95.18 & 54.36\\
    ChemT5-small & 96.22 & 13.94 & 13.50 & 96.74 & 55.10 & 89.32 & 20.89 & 13.10 & 93.47 & 54.19\\
    ChemT5-base  & 95.48 & 15.12 & 13.91 & 97.15 & 55.42 & 89.42 & 22.40 & 13.98 & 92.43 & 54.56\\
    Mol-Instruction & 65.75 & 32.01 & 26.50 & 77.91 & 50.54 & 23.40 & 37.37 & 27.97 & 71.10 & 39.96  \\
    3M-Diffusion & 87.09 & 55.36 & 34.03 & 100.0 & 69.12 & 87.05 & 64.41 & 33.44 & 100.0 & 71.22 \\
    \midrule
    \textbf{Ours-MF}&84.85&58.80&49.37&100.0&73.26&79.90&67.80&52.30&100.0&75.00\\
    \textbf{Ours-VMF}&76.54&70.77&59.78&100.0&76.77&75.23&70.99&57.29&100.0&75.88\\
    \textbf{Ours-VMFD} &77.02&69.33&58.08&100.0&76.11&78.55&68.67&54.69&100.0&75.48\\
    \textbf{Ours-MFD}&83.68&59.26&60.34&100.0&75.82&80.21&68.05&49.30&100.0&74.39\\
   
    \bottomrule
\end{tabular}
\caption{Quantitative comparison of conditional generation on ChEBI-20 and PubChem. Our method significantly outperforms SOTA baselines in novelty (Nov.), diversity (Div.), and validity (Val.), while maintaining strong similarity (Sim.) and excelling in overall (Ove.) performance. Results are percentages (higher is better).}
\label{tab:cond_1}
\end{threeparttable}

\end{table*}

\begin{table*}
\centering

\begin{threeparttable}
\setlength{\tabcolsep}{3mm}
\begin{tabular}{l | c c c c c | c c c c c}
    \toprule
    & \multicolumn{5}{c|}{\textbf{PCDes}} & \multicolumn{5}{c}{\textbf{MoMu}} \\
    Methods & Sim.& Nov.& Div.& Val.& Ove.& Sim.& Nov.& Div.& Val.& Ove.\\
    \midrule
    MolT5-small & 64.84 & 24.91 & 9.67 & 73.96 & 43.35 & 16.64 & 97.49 & 29.95 & 60.19 & 51.07 \\
    MolT5-base  &71.71 & 25.85 & 10.50 & 81.92 & 47.50 & 19.76 & 97.78 & 29.98 & 68.84 & 54.09 \\
    MolT5-large & 88.37 & 20.15 & 9.49 & 96.48 & 53.62 & 25.07 & 97.47 & 30.33 & 90.40 & 60.82 \\
    ChemT5-small & 86.27 & 23.28 & 13.17 & 93.73 & 54.11 & 23.25 & 96.97 & 30.04 & 88.45 & 59.68 \\
    ChemT5-base  & 85.01 & 25.55 & 14.08 & 92.93 & 54.39 & 23.40 & 97.65 & 30.07 & 87.61 & 59.68 \\
    Mol-Instruction & 60.86 & 35.60 & 24.57 & 79.19 & 50.06 & 14.89 & 97.52 & 30.17 & 68.32 & 52.73 \\
    3M-Diffusion & 81.57 & 63.66 & 32.39 & 100.0 & 69.41 & 24.62 & 98.16 & 37.65 & 100.0 & 65.11 \\
    \midrule
    \textbf{Ours-MF}&73.49&69.83&51.25&100.0&73.64&24.61&97.90&59.82&100.0&70.58\\
    \textbf{Ours-VMF}&72.09&74.50&61.61&100.0&77.05&24.71&97.94&70.29&100.0&73.24\\
    \textbf{Ours-VMFD}&69.00&73.86&55.25&100.0&74.53&24.71&97.49&62.50&100.0&71.18\\
    \textbf{Ours-MFD} &72.89&72.11&54.96&100.0&74.99&26.78&97.34&63.18&100.0&71.83\\
    
    \bottomrule
\end{tabular}
\caption{Quantitative comparison of conditional generation on PCDes and MoMu. Our method surpasses SOTA baselines in novelty (Nov.), diversity (Div.), and validity (Val.), while maintaining strong similarity (Sim.) and consistently leading in overall (Ove.) scores. Results are in percentages (higher is better).}
\label{tab:cond_2}
\end{threeparttable}

\end{table*}

\begin{table*}
\centering

\begin{threeparttable}
\setlength{\tabcolsep}{2mm}
\begin{tabular}{l | c c c c c c | c c c c c c}
    \toprule
    & \multicolumn{6}{c|}{\shortstack[c]{\textbf{ChEBI-20}}} & \multicolumn{6}{c}{\shortstack[c]{\textbf{PubChem}}} \\
    Methods & Uni.& Nov. &  KL & FCD& Val. & Ove. & Uni.& Nov.&  KL& FCD& Val. & Ove.\\
    \midrule
    CharRNN & 72.46 & 11.57 & 95.21 & 75.95  & 98.21 & 70.68 & 63.28 & 23.47 & 90.72 & 76.02 & 94.09 & 69.52 \\
    VAE & 57.57 & 47.88 & 95.47 & 74.19  & 63.84& 67.79 & 44.45& 42.47 & 91.67 & 55.56 & 94.10 &65.65 \\
    AAE & 1.23 & 1.23 & 38.47 & 0.06  & 1.35 & 8.47 & 2.94 & 3.21 & 39.33 & 0.08 & 1.97  & 9.51\\
    LatentGAN & 66.93 & 57.52 & 94.38 & 76.65  & 73.02& 73.70& 52.00 & 50.36 & 91.38 & 57.38 & 53.62 & 60.95 \\
    BwR & 22.09 & 21.97 & 50.59 & 0.26 & 22.66 & 23.51& 82.35 & 82.34 & 45.53 & 0.11 & 87.73 & 59.61\\ 
    HierVAE & 82.17 & 72.83 & 93.39 & 64.32 & 100.0 & 82.54& 75.33 & 72.44 & 89.05 & 50.04 & 100.0  &77.37\\ 
    PS-VAE & 76.09 & 74.55 & 83.16 & 32.44 & 100.0& 73.25& 66.97 & 66.52 & 83.41 & 14.41 & 100.0  & 66.26\\
    3M-Diffusion & 83.04 & 70.80 & 96.29 & 77.83 & 100.0& 85.59& 85.42 & 81.20 & 92.67 & 58.27 & 100.0 & 83.51\\
    \midrule
    % \textbf{Ours-MF (50 NFE)}&74.50&66.97&93.63&65.63&100.0&80.15&90.09&85.88&95.08&74.12&100.0&89.03\\
     \textbf{Ours}&80.74&71.75&94.33&76.20&100.0&84.60&91.64&87.66&92.55&61.07&100.0&86.58\\
    \bottomrule
\end{tabular}
\caption{Quantitative comparison of unconditional generation. Results of Uniq (Uni.), KL Div (KL), and FCD on ChEBI-20 and PubChem, which refer to Uniqueness, KL Divergence, and Fréchet ChemNet Distance, respectively. Results are presented in percentage values. A higher number indicates a better generation quality.}
\label{tab:uncond}
\end{threeparttable}

\end{table*}
In this section, we describe the experimental setup and summarize the evaluation results.

\paragraph{Datasets.} We conduct experiments on four molecular datasets: PubChem~\cite{liu2023molca}, ChEBI-20~\cite{edwards2021text2mol}, PCDes~\cite{zeng2022deep}, and MoMu~\cite{su2022molecular}. Following prior work on molecular structure generation~\cite{irwin2012zinc, blum2009970, rupp2012fast, zhu20243m}, we limit our analysis to molecules with fewer than 30 atoms, consistent with standard practice~\cite{ramakrishnan2014quantum, polykovskiy2020molecular, brown2019guacamol}. For instance, datasets like QM9 contain molecules with up to 9 heavy atoms~\cite{ramakrishnan2014quantum}, while ZINC-based datasets such as ZINC-250K and MOSES constrain compounds to a similar size range~\cite{irwin2012zinc, polykovskiy2020molecular}. The GuacaMol benchmark likewise targets lead-like molecules in this range~\cite{brown2019guacamol}. Dataset statistics are shown in Table~\ref{tab:stats}. Note that PCDes and MoMu share the same training/validation splits but differ in test sets, and all comparisons are reported on the respective test sets.
\begin{table}[t]
    \centering

    \begin{tabular}{lccc}
        \toprule
        \textbf{Dataset} & \textbf{Training} & \textbf{Validation} & \textbf{Test} \\
        \midrule
        ChEBI-20  & 15,409  & 1,971  & 1,965  \\
        PubChem   & 6,912   & 571    & 1,162  \\
        PCDes     & 7,474   & 1,051  & 2,136  \\
        MoMu      & 7,474   & 1,051  & 4,554  \\
        \bottomrule
    \end{tabular}
    \caption{Dataset statistics for training, validation, and test.}
    \label{tab:stats}
\end{table}

\paragraph{Implementation details.}
This section outlines our implementation for conditional and unconditional molecule generation, along with baseline models for comparison.

For instruction-guided/conditional generation, our model is benchmarked against MolT5~\cite{edwards2022translation}, ChemT5~\cite{christofidellis2023unifying}, Mol-Instruction~\cite{fang2023mol}, and 3M-Diffusion~\cite{zhu20243m}, considering multiple variants of MolT5 (small, base, large) and ChemT5 (small, base). For the unconditional scenario, we compare with prominent approaches: CharRNN~\cite{segler2018generating}, VAE~\cite{kingma2013auto}, AAE~\cite{makhzani2015adversarial}, LatentGAN~\cite{prykhodko2019novo}, BwR~\cite{diamant2023improving}, HierVAE~\cite{jin2020hierarchical}, PS-VAE~\cite{kong2022molecule}, and 3M-Diffusion~\cite{zhu20243m}. To ensure fair comparison, we use the same train, validation, and test splits as 3M-Diffusion and report its results from the original paper to avoid discrepancies from environment or hyper-parameters.

Our architecture uses GIN~\cite{hu2019strategies} as the graph encoder for molecular structures, SciBERT~\cite{beltagy2019scibert} for text encoding, and HierVAE~\cite{jin2020hierarchical} as the graph decoder. The converted latent representation is utilized via our proposed VMF training, where $\theta$ utilizes transformer architecture, maintaining causal dependencies, and $\phi$ uses MLPs. We set sampling steps to 1-NFE for conditional generation and 3/5-NFE for unconditional generation. The model is optimized using the Adam optimizer~\cite{jimmy2014adam} with a learning rate of $0.001$ and trained for 1000 epochs.

Classifier-free guidance~\cite{ho2022classifier} is implemented by randomly dropping conditional embeddings with 0.1 probability during training and inference; for unconditional generation, all conditional inputs are removed. Models are implemented in PyTorch~\cite{paszke2019pytorch} and trained on NVIDIA A100 GPUs. Code is provided in the supplementary files.

\paragraph{Performance Metrics.}
We evaluate our model on both conditional (text-guided) and unconditional molecule generation tasks. For conditional generation, we follow standard protocols~\cite{edwards2022translation, christofidellis2023unifying, fang2023mol, zhu20243m} and assess: 
(\textbf{i}) \textit{Similarity}, the proportion of generated molecules matching the ground truth with MACCS~\cite{durant2002reoptimization} cosine similarity with a threshold of $0.5$; 
(\textbf{ii}) \textit{Novelty}, the fraction of generated molecules with $f(G, \hat{G}) < 0.8$, indicating that the generated molecules differ from the references; 
(\textbf{iii}) \textit{Diversity}, defined as the average pairwise distance $1 - f(\cdot, \cdot)$ among valid molecules (where $f$ is the MACCS similarity and a molecule is considered valid if $f(G, \hat{G}) \ge 0.5$); and 
(\textbf{iv}) \textit{Validity}, the percentage of chemically valid molecules.

For unconditional generation, we rely on the GuacaMol benchmarks~\cite{brown2019guacamol} and report: 
(\textbf{i}) \textit{Uniqueness}, the ratio of distinct molecules; 
(\textbf{ii}) \textit{Novelty}, the proportion of molecules not present in the training set; 
(\textbf{iii}) \textit{KL Divergence}, which quantifies the distributional similarity between generated and training molecules; and 
(\textbf{iv}) \textit{Fréchet ChemNet Distance} (FCD)~\cite{preuer2018frechet}, which measures feature-level alignment using a ChemNet encoder. 
All metrics are normalized to [0, 1] and reported as percentages, where higher is better.

\begin{figure}[H]
    \centering
    \includegraphics[width=0.7\linewidth]{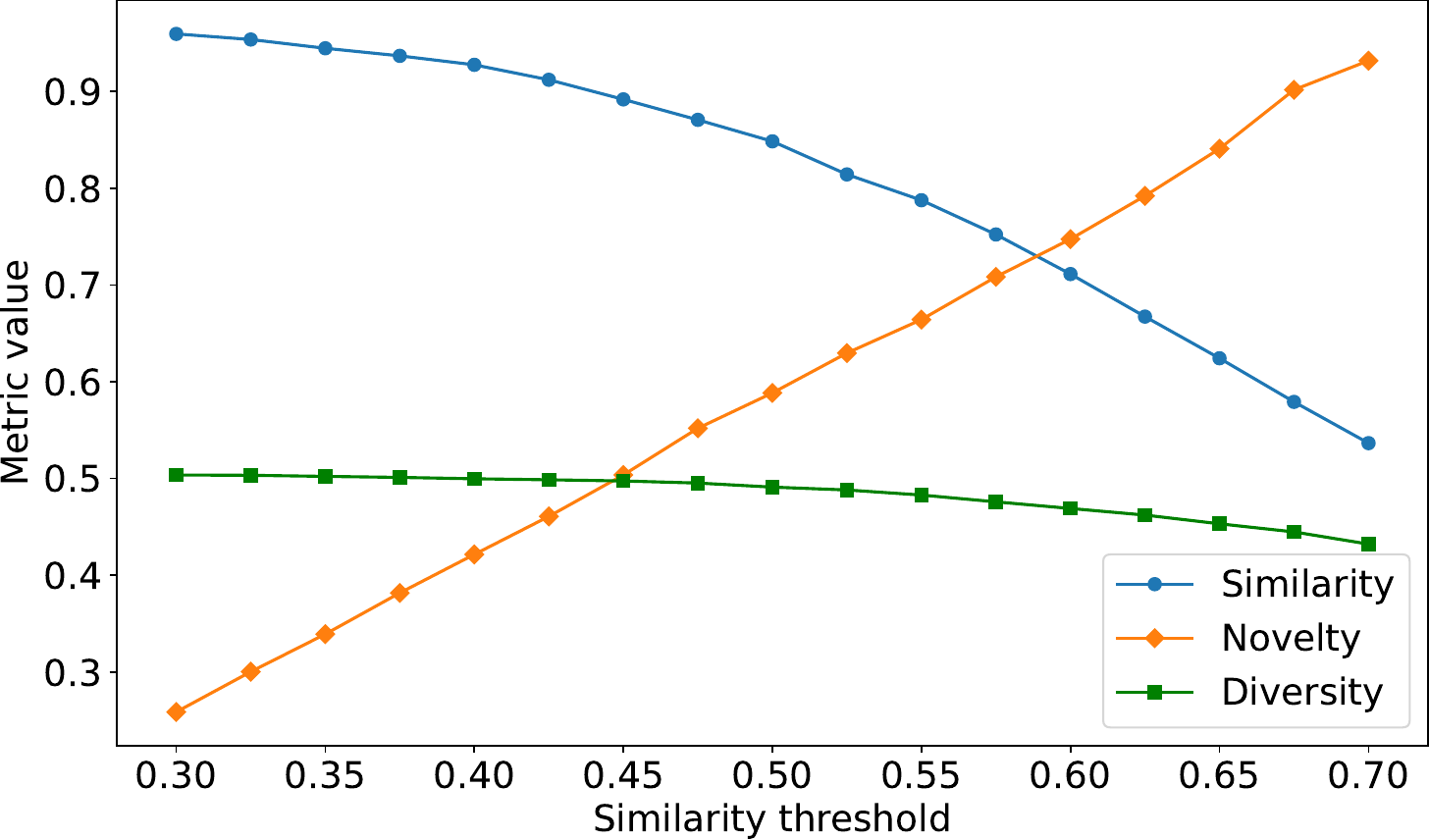}
    \caption{Tradeoff curve demonstrating result variance.}
    \label{fig:tradeoff}
\end{figure}

Individual metrics often show trade-offs. As shown in Figure~\ref{fig:tradeoff}, raising the similarity threshold decreases similarity and diversity but boosts novelty, revealing metric tension. To address this, we introduce an (\textbf{v}) \textit{Overall} metric, calculated as the average of all metrics, for balanced evaluation, preventing models from being favored for excelling in only one metric.
\paragraph{Result analysis.}
We evaluate our proposed variants: Mean Flow (MF), Variational Mean Flow (VMF), Mean Flow with Dispersive Loss (MFD), and Variational Mean Flow with Dispersive Loss (VMFD) against state-of-the-art (SOTA) models on four benchmark datasets. Results are summarized in Tables~\ref{tab:cond_1},~\ref{tab:cond_2}, and~\ref{tab:uncond}.

In the conditional generation setting (Tables~\ref{tab:cond_1} and~\ref{tab:cond_2}), our models consistently outperform baselines such as MolT5, ChemT5, and 3M-Diffusion in novelty, diversity, and validity. While MolT5-large achieves high similarity (e.g., $96.88\%$ on ChEBI-20), it suffers from low novelty and diversity due to overfitting. In contrast, VMF and VMFD maintain strong similarity while significantly improving diversity (up to $61.61\%$ on PCDes) and novelty (above $70\%$ across datasets), leading to the highest overall (Ove.) scores: $77.05\%$ on PCDes and $75.88\%$ on PubChem. MFD also highlights the benefit of dispersive loss, boosting diversity (e.g., $60.34\%$ on ChEBI-20).

For the unconditional generation task (Table~\ref{tab:uncond}), our MF variant achieves competitive or superior performance compared to both VAE-based and diffusion-based baselines. On ChEBI-20, MF reaches $84.60\%$ overall, closely matching or exceeding 3M-Diffusion, with perfect validity ($100\%$). On PubChem, MF attains $91.64\%$ uniqueness and $87.66\%$ novelty, outperforming all baselines by a notable margin. Due to space limits, visual case studies are provided in the Appendix.

Figure~\ref{fig:inference_time} compares inference times across models, showing that our method (MolSnap) is over an order of magnitude faster than 3M-Diffusion and 10--50$\times$ faster than transformer-based models like MolT5 and ChemT5. This speedup comes from our novel flow-based design, which requires far fewer function evaluations than diffusion-based methods.

These results highlight two key strengths of our approach: (\textbf{i}) the integration of variational modeling with Mean Flow enhances the exploration of chemical space, leading to higher novelty and diversity; and (\textbf{ii}) the proposed flow-based training framework enables faster sampling while maintaining similarity, with fewer function evaluations required compared to diffusion-based models.

Our framework demonstrates strong generalization across both conditional and unconditional tasks, outperforming SOTA models in most metrics and achieving the best overall balance between similarity, novelty, diversity, and validity.
\begin{figure}[t]
    \centering
    \includegraphics[width=0.7\linewidth]{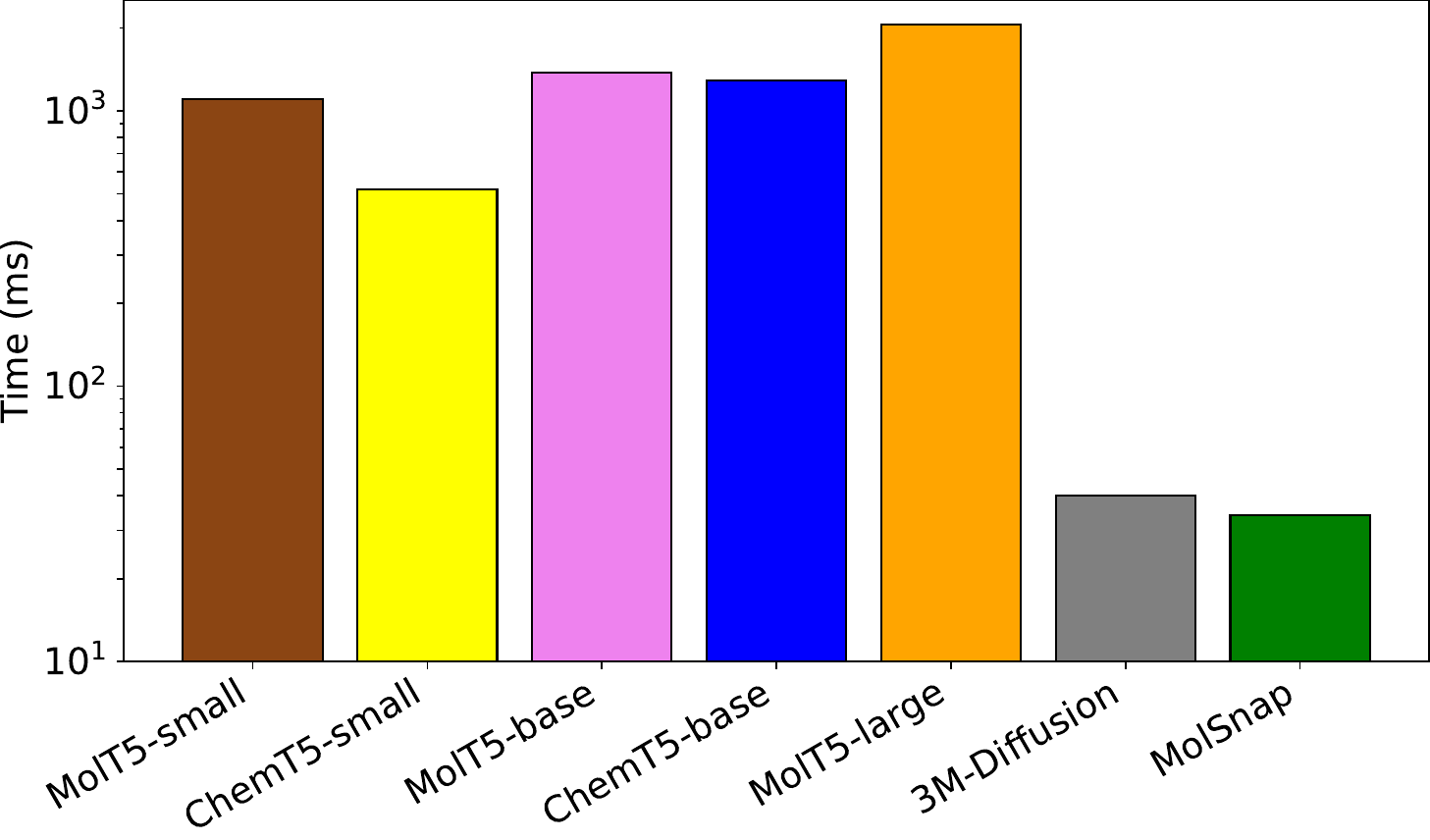}
    \caption{Inference time comparison}
    \label{fig:inference_time}
\end{figure}

\section{Ablation}
In this section, we present ablation studies on key components of our framework. While Tables~\ref{tab:cond_1} and~\ref{tab:cond_2} report results for multiple variants, we further analyze additional factors affecting performance. We report similarity, diversity, and novelty metrics, as validity remains unchanged.

\subsection{Adaptive vs. Regular $L_2$ Loss}
\begin{figure}[t]
    \centering
    \begin{subfigure}[t]{0.48\linewidth}
    
        \centering        
        \includegraphics[width=\linewidth]{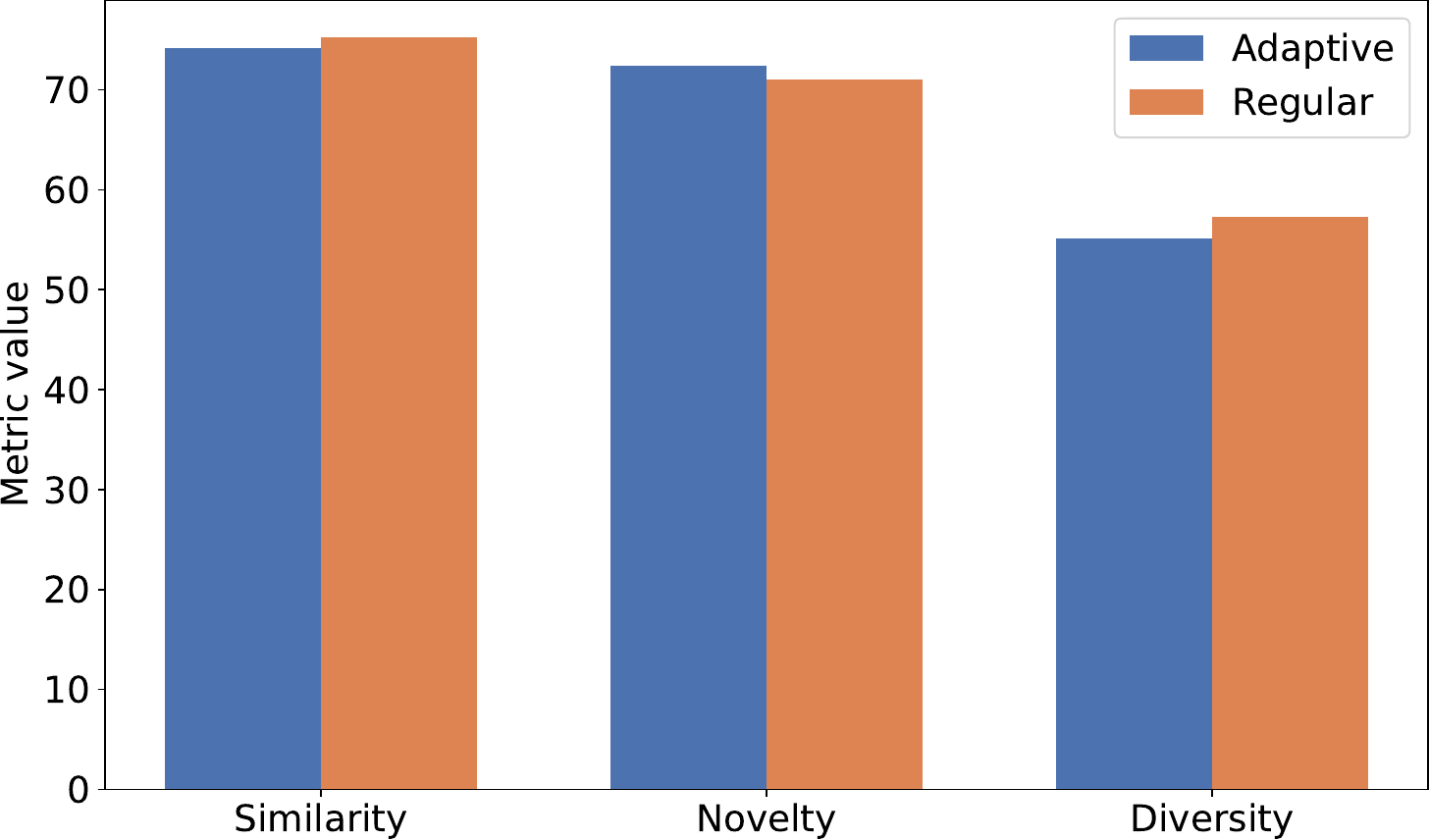}
    \caption{$L_2$ loss ablation}
    \label{fig:l2_ablation}
    \end{subfigure}
    \hfill
    \begin{subfigure}[t]{0.48\linewidth}
        \centering
        \includegraphics[width=\linewidth]{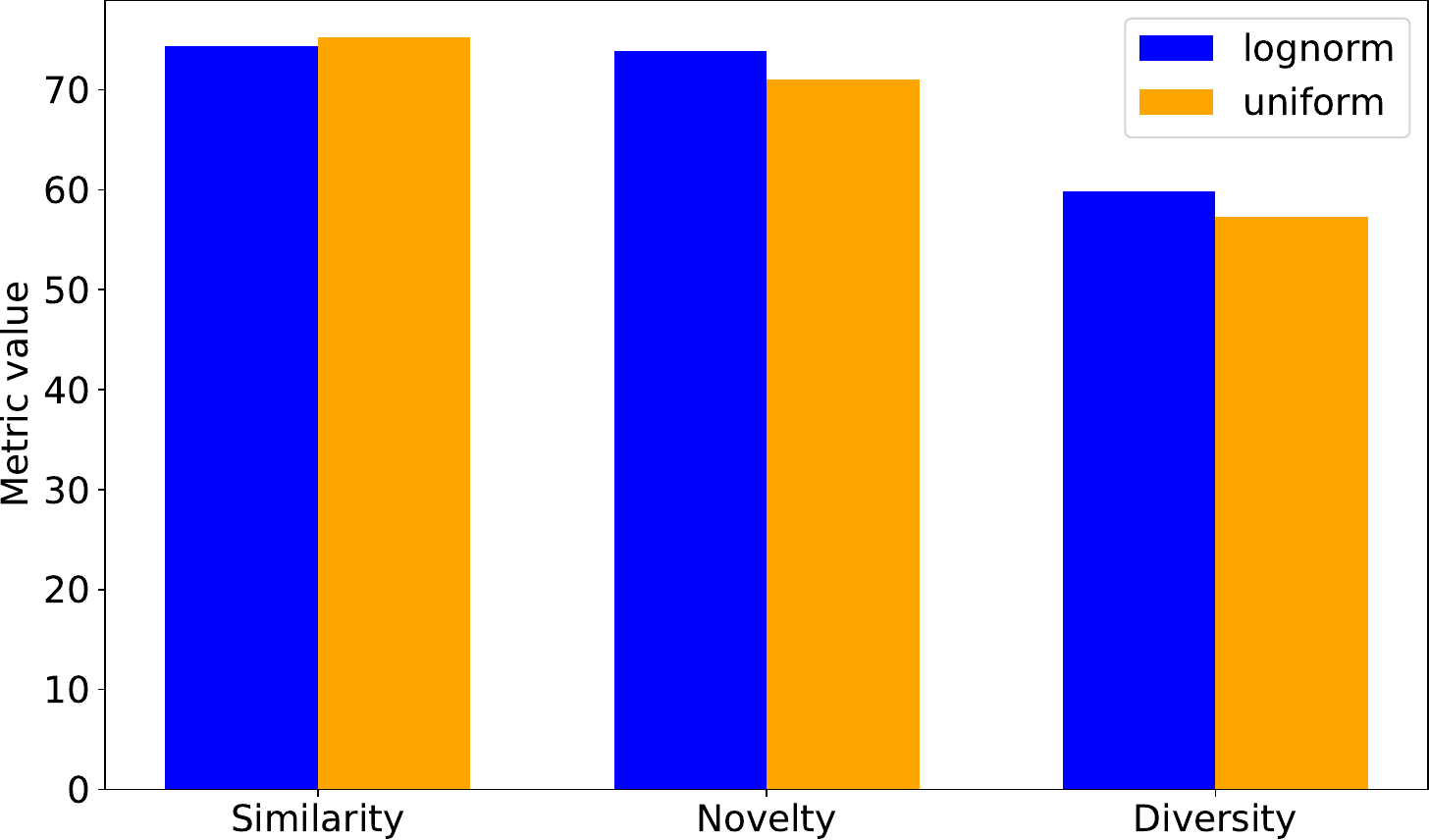}
    \caption{$t,r$ sampling strategy.}
    \label{fig:ablation_uniform_lognorm}
    \end{subfigure}
    \caption{Ablation on loss and time sampling strategy.}
    \label{fig:ablation}
    % \vspace{-20pt}
\end{figure}

Our framework uses the standard $L_2$/Mean Squared Error (MSE) loss. Mean Flow~\cite{geng2025mean} proposed an adaptive $L_2$ loss, which we incorporated to evaluate its effect. Figure~\ref{fig:l2_ablation} presents a comparison on the PubChem dataset, showing that the regular $L_2$ loss outperforms adaptive $L_2$ in similarity and diversity metrics.
% \begin{figure}[H]
%     \centering
%     \includegraphics[width=0.7\linewidth]{figures/adaptive_l2_ablation.pdf}
%     \caption{Ablattion analysis on $L_2$ loss}
%     \label{fig:l2_ablation}
% \end{figure}
\subsection{Sampling strategy for $t,r$}
We also evaluated uniform and log-normal sampling strategies for $t$ and $r$, following Mean Flow~\cite{geng2025mean}. Our experiments reveal that each strategy offers trade-offs across metrics: uniform sampling tends to improve novelty and diversity, while log-normal sampling better preserves similarity, as shown in Figure~\ref{fig:ablation_uniform_lognorm} for PubChem.
% \begin{figure}[H]
%     \centering
%     \includegraphics[width=0.7\linewidth]{figures/uniform_lognorm.pdf}
%     \caption{Sampling strategy for $t,r$ (uniform vs lognorm)}
%     \label{fig:ablation_uniform_lognorm}
% \end{figure}
\subsection{Effect of Classifier-free guidance}
\begin{figure}[H]
    \centering
    \includegraphics[width=0.7\linewidth]{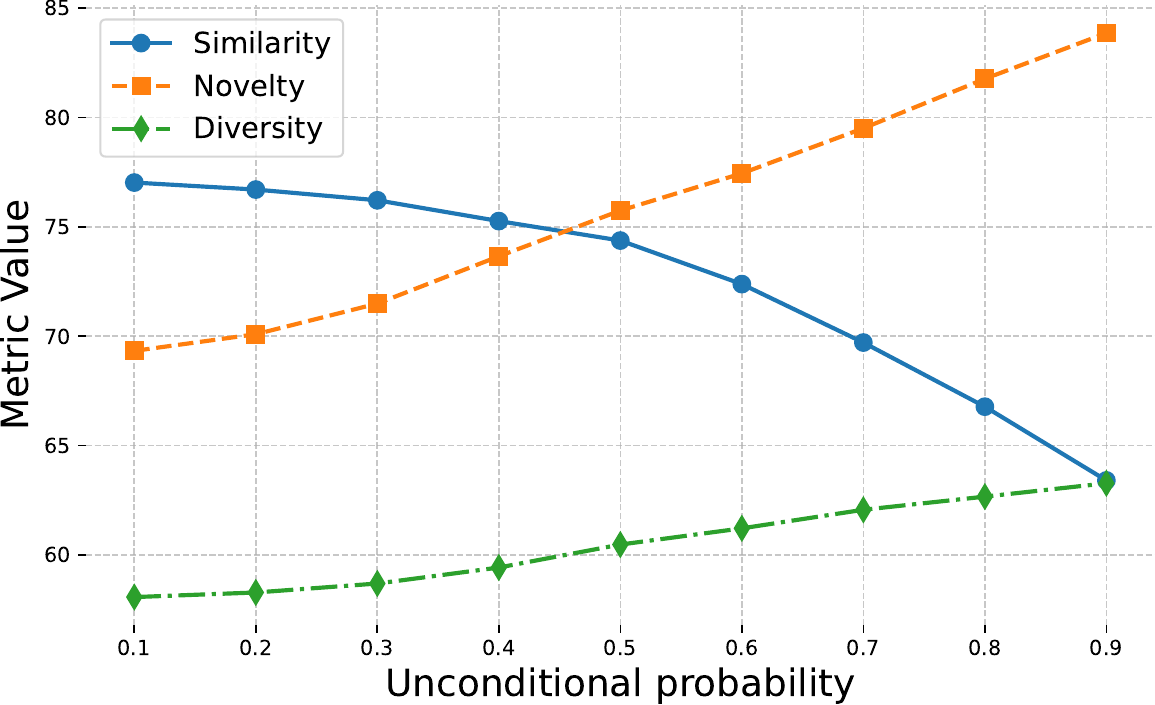}
    \caption{Performance variation with CFG.}
    \label{fig:cfg_scale_ablation}
\end{figure}

We analyze the impact of classifier-free guidance (CFG) on conditional generation. Figure~\ref{fig:cfg_scale_ablation} shows how similarity, novelty, and diversity vary with different CFG scales. Higher unconditional probability boosts novelty and diversity by encouraging chemical space exploration but reduces similarity as the model becomes less condition-aware. A moderate CFG scale offers a balanced trade-off, generating relevant and diverse molecules.

\section{Conclusion}
We proposed a novel framework for conditional molecular generation that combines a causality-aware Transformer (CAT) with a variational mean flow (VMF) approach. CAT captures the causal dependencies in molecular assembly, while VMF models the latent space as a Gaussian mixture, enabling better representation of multimodal molecular distributions. Extensive experiments on four molecular benchmarks show that our method outperforms state-of-the-art models in novelty, diversity, and validity. VMF consistently achieves 100\% validity and improves diversity (up to 70.3\%) and novelty (up to 74.5\%) while enabling fast inference with only 1–5 number of function evaluations, offering significant computational advantages over diffusion-based methods.
Our integration of causality modeling with variational flow matching offers a compelling path forward for efficient and interpretable molecular generation.

\textbf{Limitations and Future Work.} Future directions include modeling more complex molecular interactions beyond token-level causality, extending the framework to larger molecular structures, and integrating 3D structural information to enhance expressiveness for structure-based drug design.

\clearpage
\bibliography{aaai2026}

\begin{thebibliography}{59}
\providecommand{\natexlab}[1]{#1}

\bibitem[{Beltagy, Lo, and Cohan(2019)}]{beltagy2019scibert}
Beltagy, I.; Lo, K.; and Cohan, A. 2019.
\newblock SciBERT: A pretrained language model for scientific text.
\newblock \emph{arXiv preprint arXiv:1903.10676}.

\bibitem[{Bickerton et~al.(2012)Bickerton, Paolini, Besnard, Muresan, and Hopkins}]{bickerton2012quantifying}
Bickerton, G.~R.; Paolini, G.~V.; Besnard, J.; Muresan, S.; and Hopkins, A.~L. 2012.
\newblock Quantifying the chemical beauty of drugs.
\newblock \emph{Nature Chemistry}, 4(2): 90--98.

\bibitem[{Bilodeau et~al.(2022)Bilodeau, Jin, Jaakkola, Barzilay, and Jensen}]{bilodeau2022generative}
Bilodeau, C.~L.; Jin, W.; Jaakkola, T.~S.; Barzilay, R.; and Jensen, K.~F. 2022.
\newblock Generative models for molecular discovery: Recent advances and challenges.
\newblock \emph{WIREs Computational Molecular Science}, 12(4): e1608.

\bibitem[{Bjerrum and Threlfall(2017)}]{bjerrum2017molecular}
Bjerrum, E.~J.; and Threlfall, R. 2017.
\newblock Molecular generation with recurrent neural networks (RNNs).
\newblock \emph{arXiv preprint arXiv:1705.04612}.

\bibitem[{Blum and Reymond(2009)}]{blum2009970}
Blum, L.~C.; and Reymond, J.-L. 2009.
\newblock 970 million druglike small molecules for virtual screening in the chemical universe database GDB-13.
\newblock \emph{Journal of the American Chemical Society}, 131(25): 8732--8733.

\bibitem[{Born and Manica(2023)}]{born2023regression}
Born, J.; and Manica, M. 2023.
\newblock Regression Transformer enables concurrent sequence regression and generation for molecular language modelling.
\newblock \emph{Nature Machine Intelligence}, 5(4): 432--444.

\bibitem[{Brown et~al.(2019)Brown, Fiscato, Segler, and Vaucher}]{brown2019guacamol}
Brown, N.; Fiscato, M.; Segler, M.~H.; and Vaucher, A.~C. 2019.
\newblock GuacaMol: benchmarking models for de novo molecular design.
\newblock \emph{Journal of Chemical Information and Modeling}, 59(3): 1096--1108.

\bibitem[{Christofidellis et~al.(2023)Christofidellis, Giannone, Born, Winther, Laino, and Manica}]{christofidellis2023unifying}
Christofidellis, D.; Giannone, G.; Born, J.; Winther, O.; Laino, T.; and Manica, M. 2023.
\newblock Unifying molecular and textual representations via multi-task language modelling.
\newblock \emph{arXiv preprint arXiv:2301.12586}.

\bibitem[{Deng et~al.(2024)Deng, Zh, Li, Guan, and Fan}]{deng2024causal}
Deng, C.; Zh, D.; Li, K.; Guan, S.; and Fan, H. 2024.
\newblock Causal Diffusion Transformers for Generative Modeling.
\newblock \emph{arXiv preprint arXiv:2412.12095}.

\bibitem[{Diamant et~al.(2023)Diamant, Tseng, Chuang, Biancalani, and Scalia}]{diamant2023improving}
Diamant, N.~L.; Tseng, A.~M.; Chuang, K.~V.; Biancalani, T.; and Scalia, G. 2023.
\newblock Improving graph generation by restricting graph bandwidth.
\newblock In \emph{International Conference on Machine Learning}, 7939--7959. PMLR.

\bibitem[{Durant et~al.(2002)Durant, Leland, Henry, and Nourse}]{durant2002reoptimization}
Durant, J.~L.; Leland, B.~A.; Henry, D.~R.; and Nourse, J.~G. 2002.
\newblock Reoptimization of MDL keys for use in drug discovery.
\newblock \emph{Journal of Chemical Information and Computer Sciences}, 42(6): 1273--1280.

\bibitem[{Edwards et~al.(2022)Edwards, Lai, Ros, Honke, Cho, and Ji}]{edwards2022translation}
Edwards, C.; Lai, T.; Ros, K.; Honke, G.; Cho, K.; and Ji, H. 2022.
\newblock Translation between molecules and natural language.
\newblock \emph{arXiv preprint arXiv:2204.11817}.

\bibitem[{Edwards, Zhai, and Ji(2021)}]{edwards2021text2mol}
Edwards, C.; Zhai, C.; and Ji, H. 2021.
\newblock Text2mol: Cross-modal molecule retrieval with natural language queries.
\newblock In \emph{Proceedings of the 2021 Conference on Empirical Methods in Natural Language Processing}, 595--607.

\bibitem[{Fang et~al.(2023)Fang, Liang, Zhang, Liu, Huang, Chen, Fan, and Chen}]{fang2023mol}
Fang, Y.; Liang, X.; Zhang, N.; Liu, K.; Huang, R.; Chen, Z.; Fan, X.; and Chen, H. 2023.
\newblock Mol-Instructions: A Large-Scale Biomolecular Instruction Dataset for Large Language Models.
\newblock \emph{arXiv preprint arXiv:2306.08018}.

\bibitem[{Geng et~al.(2025)Geng, Deng, Bai, Kolter, and He}]{geng2025mean}
Geng, Z.; Deng, M.; Bai, X.; Kolter, J.~Z.; and He, K. 2025.
\newblock Mean flows for one-step generative modeling.
\newblock \emph{arXiv preprint arXiv:2505.13447}.

\bibitem[{G{\'o}mez-Bombarelli et~al.(2018)G{\'o}mez-Bombarelli, Wei, Duvenaud, Hern{\'a}ndez-Lobato, S{\'a}nchez-Lengeling, Sheberla, Aguilera-Iparraguirre, Hirzel, Adams, and Aspuru-Guzik}]{gomez2018automatic}
G{\'o}mez-Bombarelli, R.; Wei, J.~N.; Duvenaud, D.; Hern{\'a}ndez-Lobato, J.~M.; S{\'a}nchez-Lengeling, B.; Sheberla, D.; Aguilera-Iparraguirre, J.; Hirzel, T.~D.; Adams, R.~P.; and Aspuru-Guzik, A. 2018.
\newblock Automatic chemical design using a data-driven continuous representation of molecules.
\newblock \emph{ACS Central Science}, 4(2): 268--276.

\bibitem[{Guo and Schwing(2025)}]{guovariational}
Guo, P.; and Schwing, A. 2025.
\newblock Variational Rectified Flow Matching.
\newblock In \emph{Forty-second International Conference on Machine Learning}.

\bibitem[{Hajduk and Greer(2007)}]{hajduk2007decade}
Hajduk, P.~J.; and Greer, J. 2007.
\newblock A decade of fragment-based drug design: strategic advances and lessons learned.
\newblock \emph{Nature Reviews Drug Discovery}, 6(3): 211--219.

\bibitem[{Ho, Jain, and Abbeel(2020)}]{ho2020denoising}
Ho, J.; Jain, A.; and Abbeel, P. 2020.
\newblock Denoising diffusion probabilistic models.
\newblock \emph{Advances in Neural Information Processing Systems}, 33: 6840--6851.

\bibitem[{Ho and Salimans(2022)}]{ho2022classifier}
Ho, J.; and Salimans, T. 2022.
\newblock Classifier-free diffusion guidance.
\newblock \emph{arXiv preprint arXiv:2207.12598}.

\bibitem[{Hu et~al.(2019)Hu, Liu, Gomes, Zitnik, Liang, Pande, and Leskovec}]{hu2019strategies}
Hu, W.; Liu, B.; Gomes, J.; Zitnik, M.; Liang, P.; Pande, V.; and Leskovec, J. 2019.
\newblock Strategies for pre-training graph neural networks.
\newblock \emph{arXiv preprint arXiv:1905.12265}.

\bibitem[{Ilnicka and Schneider(2023)}]{ilnicka2023designing}
Ilnicka, A.; and Schneider, G. 2023.
\newblock Designing molecules with autoencoder networks.
\newblock \emph{Nature Computational Science}, 3(11): 922--933.

\bibitem[{Irwin et~al.(2012)Irwin, Sterling, Mysinger, Bolstad, and Coleman}]{irwin2012zinc}
Irwin, J.~J.; Sterling, T.; Mysinger, M.~M.; Bolstad, E.~S.; and Coleman, R.~G. 2012.
\newblock ZINC: a free tool to discover chemistry for biology.
\newblock \emph{Journal of Chemical Information and Modeling}, 52(7): 1757--1768.

\bibitem[{Jimmy and Diederik(2014)}]{jimmy2014adam}
Jimmy, B.; and Diederik, P. 2014.
\newblock Adam: A method for stochastic optimization.
\newblock \emph{arXiv preprint arXiv: 1412.6980}, 2014.

\bibitem[{Jin, Barzilay, and Jaakkola(2018)}]{jin2018junction}
Jin, W.; Barzilay, R.; and Jaakkola, T. 2018.
\newblock Junction tree variational autoencoder for molecular graph generation.
\newblock In \emph{International Conference on Machine Learning}, 2323--2332. PMLR.

\bibitem[{Jin, Barzilay, and Jaakkola(2020)}]{jin2020hierarchical}
Jin, W.; Barzilay, R.; and Jaakkola, T. 2020.
\newblock Hierarchical generation of molecular graphs using structural motifs.
\newblock In \emph{International Conference on Machine Learning}, 4839--4848. PMLR.

\bibitem[{Kingma and Welling(2013)}]{kingma2013auto}
Kingma, D.~P.; and Welling, M. 2013.
\newblock Auto-encoding variational bayes.
\newblock \emph{arXiv preprint arXiv:1312.6114}.

\bibitem[{Kong et~al.(2022)Kong, Huang, Tan, and Liu}]{kong2022molecule}
Kong, X.; Huang, W.; Tan, Z.; and Liu, Y. 2022.
\newblock Molecule generation by principal subgraph mining and assembling.
\newblock \emph{Advances in Neural Information Processing Systems}, 35: 2550--2563.

\bibitem[{Kusner, Paige, and Hern{\'a}ndez-Lobato(2017)}]{kusner2017grammar}
Kusner, M.~J.; Paige, B.; and Hern{\'a}ndez-Lobato, J.~M. 2017.
\newblock Grammar variational autoencoder.
\newblock In \emph{International Conference on Machine Learning}, 1945--1954. PMLR.

\bibitem[{Li et~al.(2018)Li, Vinyals, Dyer, Pascanu, and Battaglia}]{li2018learning}
Li, Y.; Vinyals, O.; Dyer, C.; Pascanu, R.; and Battaglia, P. 2018.
\newblock Learning deep generative models of graphs.
\newblock \emph{arXiv preprint arXiv:1803.03324}.

\bibitem[{Lipman et~al.(2022)Lipman, Chen, Ben‑Hamu, Nickel, and Le}]{lipman2022flow}
Lipman, Y.; Chen, R. T.~Q.; Ben‑Hamu, H.; Nickel, M.; and Le, M. 2022.
\newblock Flow Matching for Generative Modeling.
\newblock \emph{arXiv preprint arXiv:2210.02747}.
\newblock Presented at ICLR 2023.

\bibitem[{Liu et~al.(2018)Liu, Allamanis, Brockschmidt, and Gaunt}]{liu2018constrained}
Liu, Q.; Allamanis, M.; Brockschmidt, M.; and Gaunt, A. 2018.
\newblock Constrained graph variational autoencoders for molecule design.
\newblock \emph{Advances in Neural Information Processing Systems}, 31.

\bibitem[{Liu, Gong, and Liu(2022)}]{liu2022rectified}
Liu, X.; Gong, C.; and Liu, Q. 2022.
\newblock Flow Straight and Fast: Learning to Generate and Transfer Data with Rectified Flow.
\newblock \emph{arXiv preprint arXiv:2209.03003}.
\newblock Introduces “Rectified Flow” for straight‑path transport mappings in generative modeling.

\bibitem[{Liu et~al.(2023)Liu, Li, Luo, Fei, Cao, Kawaguchi, Wang, and Chua}]{liu2023molca}
Liu, Z.; Li, S.; Luo, Y.; Fei, H.; Cao, Y.; Kawaguchi, K.; Wang, X.; and Chua, T.-S. 2023.
\newblock Molca: Molecular graph-language modeling with cross-modal projector and uni-modal adapter.
\newblock \emph{arXiv preprint arXiv:2310.12798}.

\bibitem[{Luo, Yan, and Ji(2021)}]{luo2021graphdf}
Luo, Y.; Yan, K.; and Ji, S. 2021.
\newblock Graphdf: A discrete flow model for molecular graph generation.
\newblock In \emph{International Conference on Machine Learning}, 7192--7203. PMLR.

\bibitem[{Ma et~al.(2021)Ma, Xin, Yang, Shi, Zhang, Wang, Wang, Liu, Chu, and Fu}]{ma2021paving}
Ma, Y.-S.; Xin, R.; Yang, X.-L.; Shi, Y.; Zhang, D.-D.; Wang, H.-M.; Wang, P.-Y.; Liu, J.-B.; Chu, K.-J.; and Fu, D. 2021.
\newblock Paving the way for small-molecule drug discovery.
\newblock \emph{American Journal of Rranslational Research}, 13(3): 853.

\bibitem[{Madhawa et~al.(2019)Madhawa, Ishiguro, Nakago, and Abe}]{madhawa2019graphnvp}
Madhawa, K.; Ishiguro, K.; Nakago, K.; and Abe, M. 2019.
\newblock Graphnvp: An invertible flow model for generating molecular graphs.
\newblock \emph{arXiv preprint arXiv:1905.11600}.

\bibitem[{Makhzani et~al.(2015)Makhzani, Shlens, Jaitly, Goodfellow, and Frey}]{makhzani2015adversarial}
Makhzani, A.; Shlens, J.; Jaitly, N.; Goodfellow, I.; and Frey, B. 2015.
\newblock Adversarial autoencoders.
\newblock \emph{arXiv preprint arXiv:1511.05644}.

\bibitem[{Mandal, Mandal et~al.(2009)}]{mandal2009rational}
Mandal, S.; Mandal, S.~K.; et~al. 2009.
\newblock Rational drug design.
\newblock \emph{European Journal of Pharmacology}, 90--100.

\bibitem[{Paszke et~al.(2019)Paszke, Gross, Massa, Lerer, Bradbury, Chanan, Killeen, Lin, Gimelshein, Antiga et~al.}]{paszke2019pytorch}
Paszke, A.; Gross, S.; Massa, F.; Lerer, A.; Bradbury, J.; Chanan, G.; Killeen, T.; Lin, Z.; Gimelshein, N.; Antiga, L.; et~al. 2019.
\newblock Pytorch: An imperative style, high-performance deep learning library.
\newblock \emph{Advances in Neural Information Processing Systems}, 32.

\bibitem[{Polykovskiy et~al.(2020)Polykovskiy, Zhebrak, Sanchez-Lengeling, Golovanov, Tatanov, Belyaev, Kurbanov, Artamonov, Aladinskiy, Veselov et~al.}]{polykovskiy2020molecular}
Polykovskiy, D.; Zhebrak, A.; Sanchez-Lengeling, B.; Golovanov, S.; Tatanov, O.; Belyaev, S.; Kurbanov, R.; Artamonov, A.; Aladinskiy, V.; Veselov, M.; et~al. 2020.
\newblock Molecular sets (MOSES): a benchmarking platform for molecular generation models.
\newblock \emph{Frontiers in Pharmacology}, 11: 565644.

\bibitem[{Preuer et~al.(2018)Preuer, Renz, Unterthiner, Hochreiter, and Klambauer}]{preuer2018frechet}
Preuer, K.; Renz, P.; Unterthiner, T.; Hochreiter, S.; and Klambauer, G. 2018.
\newblock Fr{\'e}chet ChemNet distance: a metric for generative models for molecules in drug discovery.
\newblock \emph{Journal of Chemical Information and Modeling}, 58(9): 1736--1741.

\bibitem[{Prykhodko et~al.(2019)Prykhodko, Johansson, Kotsias, Ar{\'u}s-Pous, Bjerrum, Engkvist, and Chen}]{prykhodko2019novo}
Prykhodko, O.; Johansson, S.~V.; Kotsias, P.-C.; Ar{\'u}s-Pous, J.; Bjerrum, E.~J.; Engkvist, O.; and Chen, H. 2019.
\newblock A de novo molecular generation method using latent vector based generative adversarial network.
\newblock \emph{Journal of Cheminformatics}, 11: 1--13.

\bibitem[{Pyzer-Knapp et~al.(2015)Pyzer-Knapp, Suh, G{\'o}mez-Bombarelli, Aguilera-Iparraguirre, and Aspuru-Guzik}]{pyzer2015high}
Pyzer-Knapp, E.~O.; Suh, C.; G{\'o}mez-Bombarelli, R.; Aguilera-Iparraguirre, J.; and Aspuru-Guzik, A. 2015.
\newblock What is high-throughput virtual screening? A perspective from organic materials discovery.
\newblock \emph{Annual Review of Materials Research}, 45: 195--216.

\bibitem[{Ramakrishnan et~al.(2014)Ramakrishnan, Dral, Rupp, and Von~Lilienfeld}]{ramakrishnan2014quantum}
Ramakrishnan, R.; Dral, P.~O.; Rupp, M.; and Von~Lilienfeld, O.~A. 2014.
\newblock Quantum chemistry structures and properties of 134 kilo molecules.
\newblock \emph{Scientific Data}, 1(1): 1--7.

\bibitem[{Rupp et~al.(2012)Rupp, Tkatchenko, M{\"u}ller, and Von~Lilienfeld}]{rupp2012fast}
Rupp, M.; Tkatchenko, A.; M{\"u}ller, K.-R.; and Von~Lilienfeld, O.~A. 2012.
\newblock Fast and accurate modeling of molecular atomization energies with machine learning.
\newblock \emph{Physical Review Letters}, 108(5): 058301.

\bibitem[{Schneuing et~al.(2024)Schneuing, Harris, Du, Didi, Jamasb, Igashov, Du, Gomes, Blundell, Lio, Welling, Bronstein, and Correia}]{schneuing2024structure}
Schneuing, A.; Harris, C.; Du, Y.; Didi, K.; Jamasb, A.; Igashov, I.; Du, W.; Gomes, C.; Blundell, T.~L.; Lio, P.; Welling, M.; Bronstein, M.; and Correia, B. 2024.
\newblock Structure-based drug design with equivariant diffusion models.
\newblock \emph{Nature Computational Science}, 4(12): 899--909.

\bibitem[{Schwaller et~al.(2019)Schwaller, Laino, Gaudin, Bolgar, Hunter, Bekas, and Lee}]{schwaller2019molecular}
Schwaller, P.; Laino, T.; Gaudin, T.; Bolgar, P.; Hunter, C.~A.; Bekas, C.; and Lee, A.~A. 2019.
\newblock Molecular transformer: a model for uncertainty-calibrated chemical reaction prediction.
\newblock \emph{ACS Central Science}, 5(9): 1572--1583.

\bibitem[{Segler et~al.(2018)Segler, Kogej, Tyrchan, and Waller}]{segler2018generating}
Segler, M.~H.; Kogej, T.; Tyrchan, C.; and Waller, M.~P. 2018.
\newblock Generating focused molecule libraries for drug discovery with recurrent neural networks.
\newblock \emph{ACS Central Science}, 4(1): 120--131.

\bibitem[{Southey and Brunavs(2023)}]{southey2023introduction}
Southey, M.~W.; and Brunavs, M. 2023.
\newblock Introduction to small molecule drug discovery and preclinical development.
\newblock \emph{Frontiers in Drug Discovery}, 3: 1314077.

\bibitem[{Su et~al.(2022)Su, Du, Yang, Zhou, Li, Rao, Sun, Lu, and Wen}]{su2022molecular}
Su, B.; Du, D.; Yang, Z.; Zhou, Y.; Li, J.; Rao, A.; Sun, H.; Lu, Z.; and Wen, J.-R. 2022.
\newblock A molecular multimodal foundation model associating molecule graphs with natural language.
\newblock \emph{arXiv preprint arXiv:2209.05481}.

\bibitem[{Wang and He(2025)}]{wang2025diffuse}
Wang, R.; and He, K. 2025.
\newblock Diffuse and Disperse: Image Generation with Representation Regularization.
\newblock \emph{arXiv preprint arXiv:2506.09027}.

\bibitem[{Weininger(1988)}]{weininger1988smiles}
Weininger, D. 1988.
\newblock SMILES, a chemical language and information system. 1. Introduction to methodology and encoding rules.
\newblock \emph{Journal of Chemical Information and Computer Sciences}, 28(1): 31--36.

\bibitem[{Weiss et~al.(2023)Weiss, Yanes, Chakraborty, Cosmo, Bronstein, and Gershoni-Poranne}]{weiss2023guided}
Weiss, T.; Yanes, E.~M.; Chakraborty, S.; Cosmo, L.; Bronstein, A.~M.; and Gershoni-Poranne, R. 2023.
\newblock Guided diffusion for inverse molecular design.
\newblock \emph{Nature Computational Science}, 3(10): 873--882.

\bibitem[{Westermayr et~al.(2023)Westermayr, Gilkes, Barrett, and Maurer}]{westermayr2023high}
Westermayr, J.; Gilkes, J.; Barrett, R.; and Maurer, R.~J. 2023.
\newblock High-throughput property-driven generative design of functional organic molecules.
\newblock \emph{Nature Computational Science}, 3(2): 139--148.

\bibitem[{You et~al.(2018)You, Liu, Ying, Pande, and Leskovec}]{you2018graph}
You, J.; Liu, B.; Ying, Z.; Pande, V.; and Leskovec, J. 2018.
\newblock Graph convolutional policy network for goal-directed molecular graph generation.
\newblock \emph{Advances in Neural Information Processing Systems}, 31.

\bibitem[{Zang and Wang(2020)}]{zang2020moflow}
Zang, C.; and Wang, F. 2020.
\newblock Moflow: an invertible flow model for generating molecuclar graphs.
\newblock In \emph{Proceedings of the 26th ACM SIGKDD International Conference on Knowledge Discovery \& Data Mining}, 617--626.

\bibitem[{Zeng et~al.(2022)Zeng, Yao, Liu, and Sun}]{zeng2022deep}
Zeng, Z.; Yao, Y.; Liu, Z.; and Sun, M. 2022.
\newblock A deep-learning system bridging molecule structure and biomedical text with comprehension comparable to human professionals.
\newblock \emph{Nature Communications}, 13(1): 862.

\bibitem[{Zhu, Xiao, and Honavar(2024)}]{zhu20243m}
Zhu, H.; Xiao, T.; and Honavar, V.~G. 2024.
\newblock 3M-Diffusion: Latent Multi-Modal Diffusion for Language-Guided Molecular Structure Generation.
\newblock In \emph{First Conference on Language Modeling}.

\end{thebibliography}
\clearpage
\appendix
\textbf{\large APPENDIX}
\setcounter{secnumdepth}{1}

\section{Attention mask formulation}

\begin{algorithm}[H]
\caption{Split Integer with Exponential Decay}
\label{alg:split}
\begin{algorithmic}[1]
    \STATE \textbf{Input:} $S$ (Sample length), $\alpha$ (decay factor)
    \STATE \textbf{Output:} \texttt{result} (split sizes), \texttt{cumsum} (cumulative steps)
    \IF{$\alpha = 1.0$}
        \STATE $N \sim \texttt{UniformInt}(1, S)$
    \ELSE
        \STATE $\texttt{base} \gets \frac{1 - \alpha}{1 - \alpha^{S}}$
        \STATE $p[i] \gets \texttt{base} \cdot \alpha^{i}, \quad i = 0 \dots S-1$
        \STATE $N \sim \texttt{Categorical}(1 \dots S, p)$
    \ENDIF
    \STATE $\texttt{cumsum} \gets [0] \cup \texttt{sorted}(\texttt{Sample}(1 \dots S-1, N-1)) \cup [S]$
    \STATE $\texttt{result}[i] \gets \texttt{cumsum}[i+1] - \texttt{cumsum}[i], \quad i=0 \dots len(cumsum)-2$
    \RETURN \texttt{result}, \texttt{cumsum}
\end{algorithmic}
\end{algorithm}

\begin{figure}
    \centering
    \includegraphics[width=\linewidth]{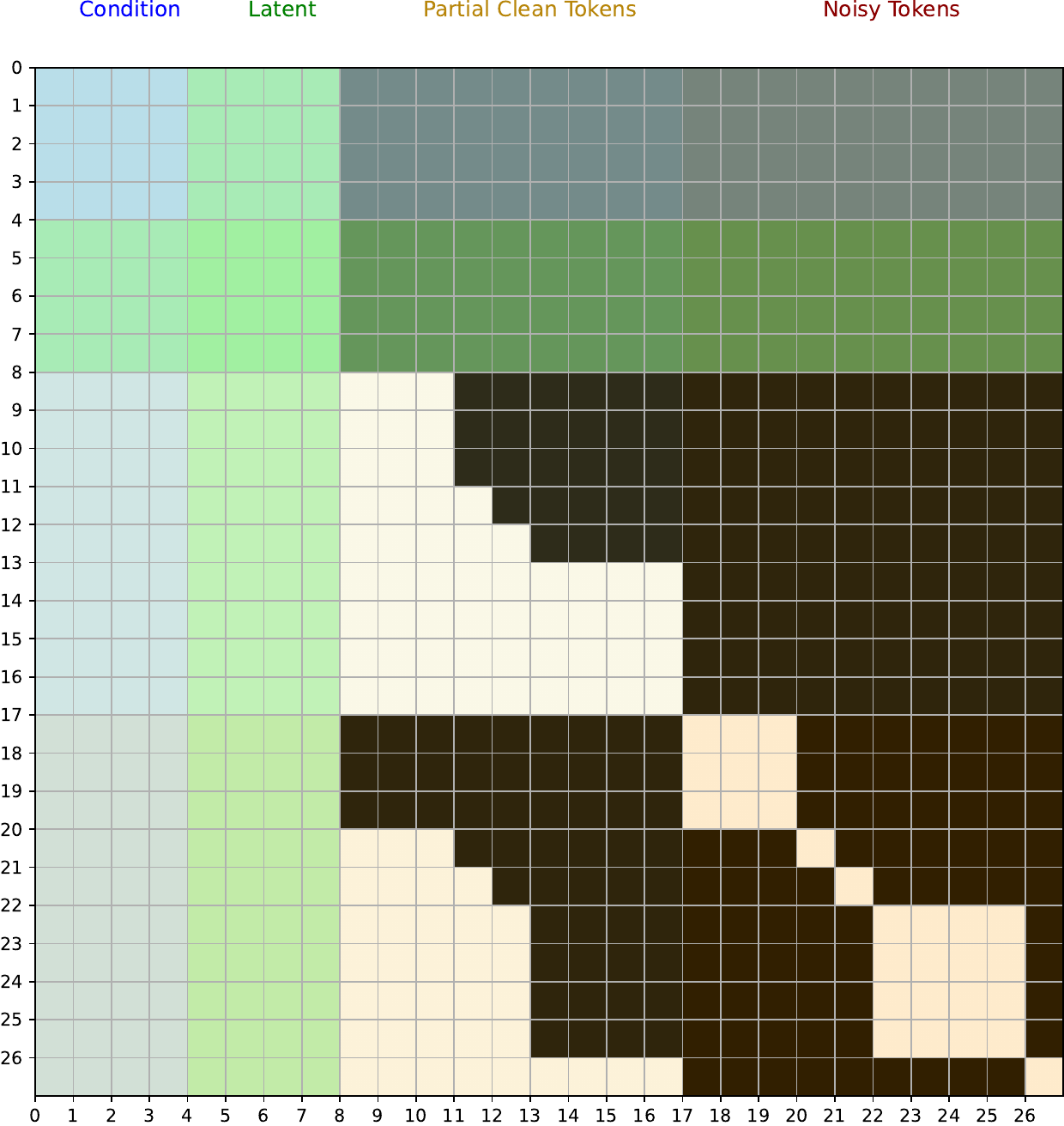}
    \caption{Example of an attention mask with token \textbf{lengths}: condition tokens ($c$) = 4, latent tokens ($h$) = 4, partial clean tokens ($x_p$) = 9, and noisy tokens ($z$) = 10. Black shaded regions indicate blocked attention, while the lighter cells represent allowable attention.}
    \label{fig:attention_mask_molsnap}
\end{figure}

\begin{algorithm}[H]
\caption{Construct Attention Mask, $m$}
\label{alg:attn_mask}
\begin{algorithmic}[1]
    \STATE \textbf{Input:} \texttt{sample\_len}, \texttt{cond\_len}, \texttt{latent\_len}, \texttt{split\_sizes}, \texttt{cumsum}
    \STATE \textbf{Output:} Attention mask $\mathbf{m}$
    \STATE $\texttt{cond\_len} \gets \texttt{cond\_len} + \texttt{latent\_len}$
    \STATE $\texttt{visible\_len} \gets \texttt{sample\_len} - \texttt{split\_sizes}[-1]$
    \STATE $\texttt{ctx\_len} \gets \texttt{cond\_len} + \texttt{visible\_len}$
    \STATE $\texttt{seq\_len} \gets \texttt{ctx\_len} + \texttt{sample\_len}$
    \STATE Initialize $\mathbf{m} \gets \mathbf{1}_{\texttt{seq\_len} \times \texttt{seq\_len}}$
    \STATE $\mathbf{m}[:, :\texttt{cond\_len}] \gets 0$
    \STATE Initialize $\mathbf{T_1}, \mathbf{T_2}, \mathbf{T_3}$ as all-ones triangular matrices
    \FOR{$i = 0$ to $|\texttt{split\_sizes}|-2$}
        \STATE $\mathbf{T_1}[\texttt{cumsum}[i]:\texttt{cumsum}[i+1], \; 0:\texttt{cumsum}[i+1]] \gets 0$
        \STATE $\mathbf{T_2}[\texttt{cumsum}[i+1]:\texttt{cumsum}[i+2], \; 0:\texttt{cumsum}[i+1]] \gets 0$
    \ENDFOR
    \FOR{$i = 0$ to $|\texttt{split\_sizes}|-1$}
        \STATE $\mathbf{T_3}[\texttt{cumsum}[i]:\texttt{cumsum}[i+1], \; \texttt{cumsum}[i]:\texttt{cumsum}[i+1]] \gets 0$
    \ENDFOR
    \STATE $\mathbf{m}[\texttt{cond\_len}:\texttt{ctx\_len}, \texttt{cond\_len}:\texttt{ctx\_len}] \gets \mathbf{T_1}$
    \STATE $\mathbf{m}[\texttt{ctx\_len}:, \texttt{cond\_len}:\texttt{ctx\_len}] \gets \mathbf{T_2}$
    \STATE $\mathbf{m}[\texttt{ctx\_len}:, \texttt{ctx\_len}:] \gets \mathbf{T_3}$
    \RETURN $\mathbf{m}[\texttt{None}, \texttt{None}, :, :]$
\end{algorithmic}
\end{algorithm}
We construct the attention mask $m$ in two steps, using Algorithm~\ref{alg:split} and Algorithm~\ref{alg:attn_mask}. First, Algorithm~\ref{alg:split} generates the autoregressive step structure by returning \texttt{split\_sizes} and their cumulative indices (\texttt{cumsum}). These are then used by Algorithm~\ref{alg:attn_mask} to construct the final attention mask $m$, which enforces directional and grouped attention among visible and noisy tokens, while allowing free attention from conditional and latent tokens. An illustration of a resulting attention mask is shown in Figure~\ref{fig:attention_mask_molsnap}.
\section{Conditional samples}
\begin{figure*}[ht]
    \centering
    
    \includegraphics[width=\linewidth]{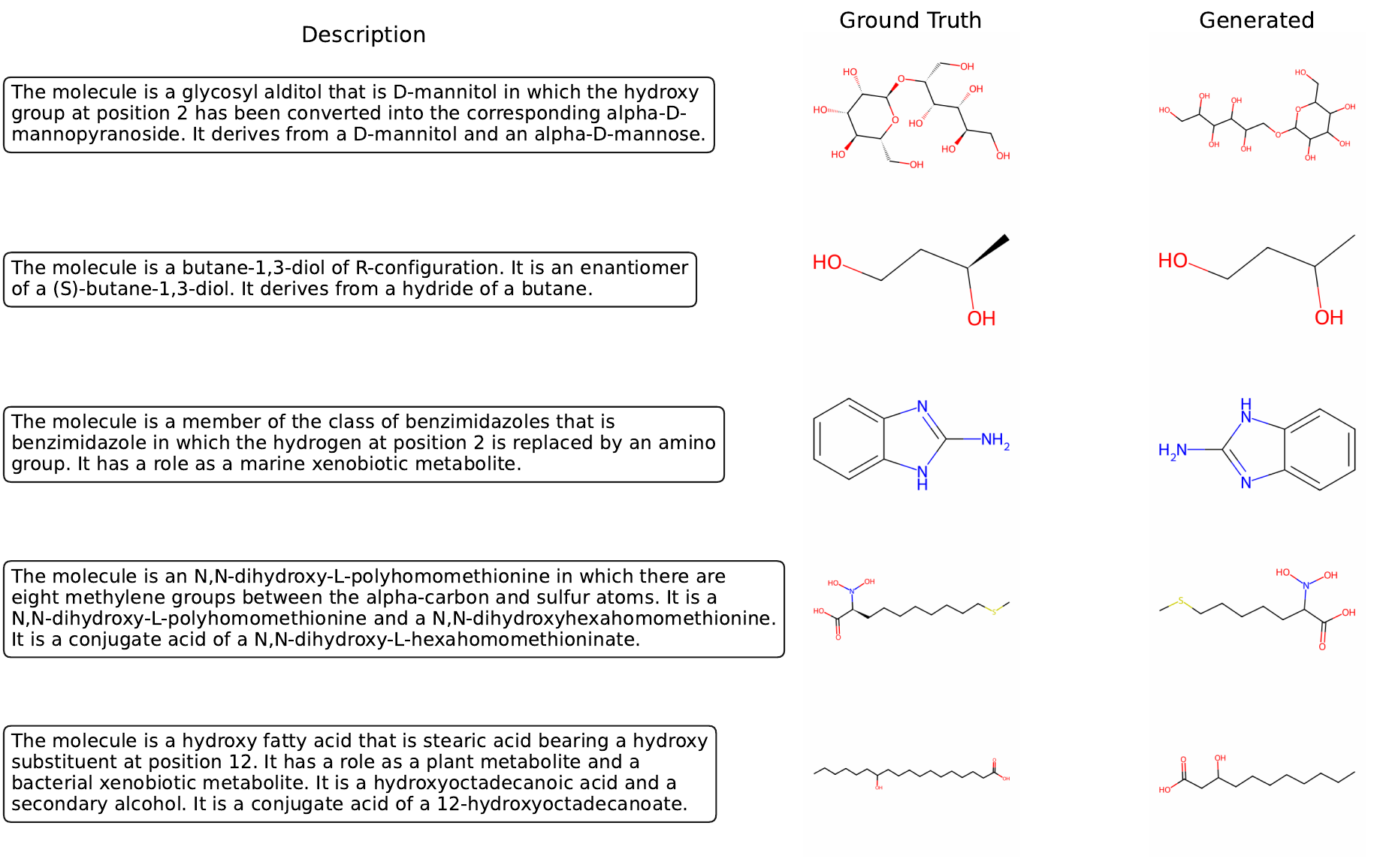}
    \caption{Text conditional molecule generation from ChEBI-20 dataset.}
    \label{fig:cond_samples}
\end{figure*}

Figure~\ref{fig:cond_samples} presents visual results of text-conditional molecule generation on the ChEBI-20 dataset. Given detailed molecular descriptions, our model generates chemically meaningful structures that closely align with the input text. The examples include glycosides, chiral diols (enantiomers), substituted benzimidazoles, methionine derivatives with long aliphatic chains, and hydroxylated fatty acids. In all cases, the generated structures capture key functional groups and molecular backbones, demonstrating strong alignment with the semantics of the input descriptions.

\begin{table*}[t]
    \centering

    \begin{tabular}{lccccc}
        & \multicolumn{5}{c}{\textbf{Condition:} The molecule is an anthocyanidin cation.}\\
         \rotatebox{90}{MolSnap}&  \includegraphics[width=0.15\linewidth,trim={8cm 14cm 43cm 2.9cm}, clip]{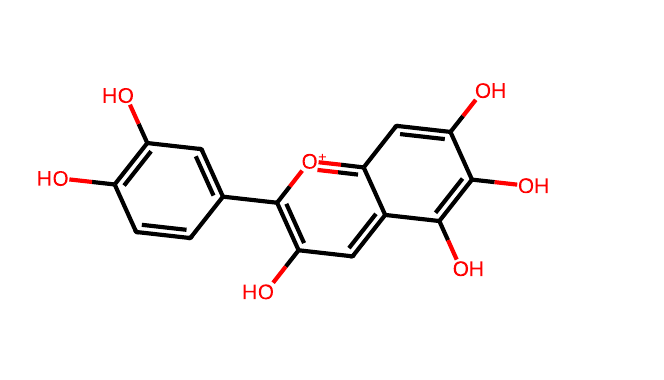}&  \includegraphics[width=0.15\linewidth,trim={0cm 0.8cm 0cm 1.1cm}, clip]{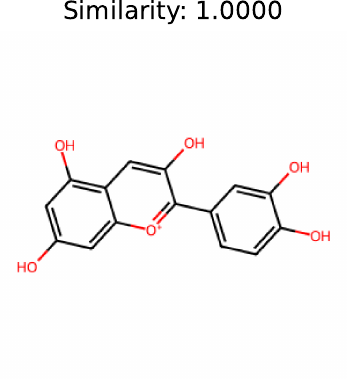}&  \includegraphics[width=0.15\linewidth,trim={0cm 1cm 0cm 1.1cm}, clip]{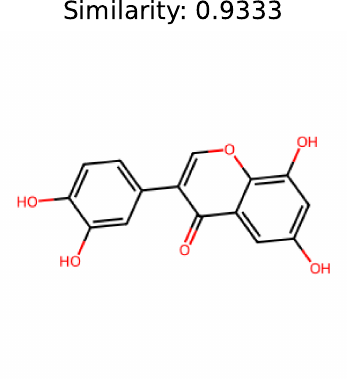}&  \includegraphics[width=0.15\linewidth,trim={0cm 0.8cm 0cm 1.1cm}, clip]{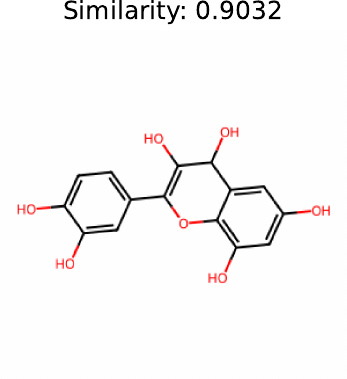}&  \includegraphics[width=0.15\linewidth,trim={0cm 0.8cm 0cm 1.1cm}, clip]{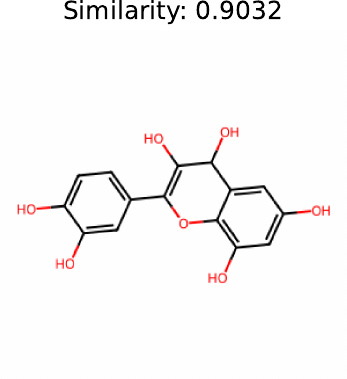}\\
         & Reference & Sim: 1.0& Sim: 0.93 & Sim: 0.90 & Sim: 0.90\\
        \rotatebox{90}{MolT5-large}&  \includegraphics[width=0.15\linewidth,trim={8cm 14cm 43cm 2.9cm}, clip]{figures/case_an.pdf} & \includegraphics[width=0.15\linewidth,trim={18cm 14cm 33cm 2.5cm}, clip]{figures/case_an.pdf} & \includegraphics[width=0.15\linewidth,trim={28cm 14cm 23cm 2.5cm}, clip]{figures/case_an.pdf} & \includegraphics[width=0.15\linewidth,trim={40cm 14cm 11cm 2.5cm}, clip]{figures/case_an.pdf} & \includegraphics[width=0.15\linewidth,trim={52cm 14cm 1cm 2.5cm}, clip]{figures/case_an.pdf}\\
        & Reference &  Sim: 0.74 & Sim: 0.74 & Sim: 0.74 & Sim: 0.73\\
        \rotatebox{90}{3M-Diffusion}&   \includegraphics[width=0.15\linewidth,trim={8cm 14cm 43cm 2.9cm}, clip]{figures/case_an.pdf} & \includegraphics[width=0.15\linewidth,trim={18cm 2.5cm 33cm 14cm}, clip]{figures/case_an.pdf} & \includegraphics[width=0.15\linewidth,trim={28cm 2.5cm 23cm 14cm}, clip]{figures/case_an.pdf} & \includegraphics[width=0.15\linewidth,trim={40cm 2.5cm 11cm 14cm}, clip]{figures/case_an.pdf} & \includegraphics[width=0.15\linewidth,trim={52cm 2.5cm 1cm 14cm}, clip]{figures/case_an.pdf}\\
        & Reference &  Sim: 0.87 & Sim: 0.85 & Sim: 0.83 & Sim: 0.81\\
          
    \end{tabular}
    \caption{Comparison of molecules generated by MolSnap, 3M-Diffusion, and MolT5-large under the anthocyanidin cation condition. Higher similarity (Sim.) denotes better alignment with the reference.}
    \label{tab:case_an}
\end{table*}
\begin{table*}[t]
    \centering
    \resizebox{\linewidth}{!}{
    \begin{tabular}{lccccc}
        & \multicolumn{5}{c}{\textbf{Condition:} This molecule is like a drug.}\\
         \rotatebox{90}{MolSnap}&  \includegraphics[width=0.15\linewidth,trim={0cm 0.8cm 0cm 1.1cm}, clip]{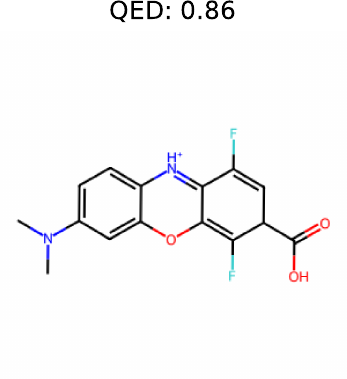}&  \includegraphics[width=0.15\linewidth,trim={0cm 0.8cm 0cm 1.1cm}, clip]{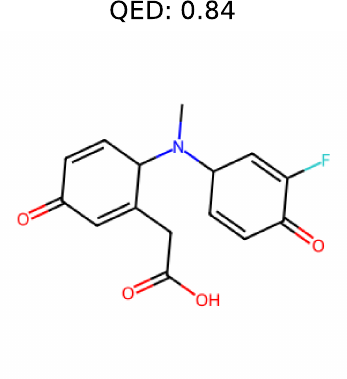}&  \includegraphics[width=0.15\linewidth,trim={0cm 1cm 0cm 1.1cm}, clip]{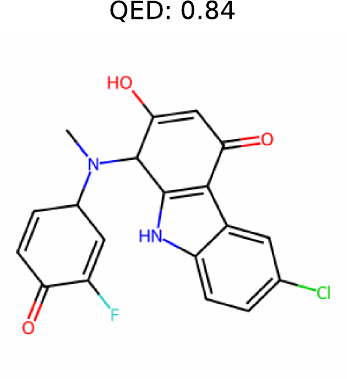}&  \includegraphics[width=0.15\linewidth,trim={0cm 0.8cm 0cm 1.1cm}, clip]{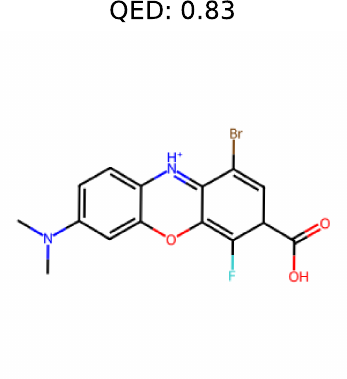}&  \includegraphics[width=0.15\linewidth,trim={0cm 0.8cm 0cm 1.1cm}, clip]{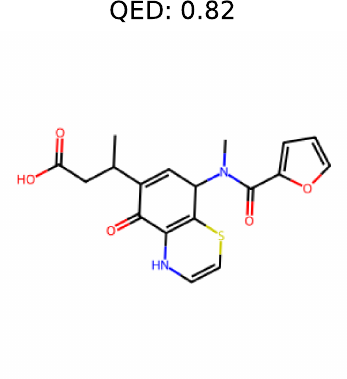}\\
         & QED: 0.86 & QED: 0.84 & QED: 0.84 & QED: 0.83 & QED: 0.82\\
        \rotatebox{90}{MolT5-large}&  \rotatebox{90}{\includegraphics[width=0.15\linewidth,trim={11cm 12cm 44cm 2.9cm}, clip]{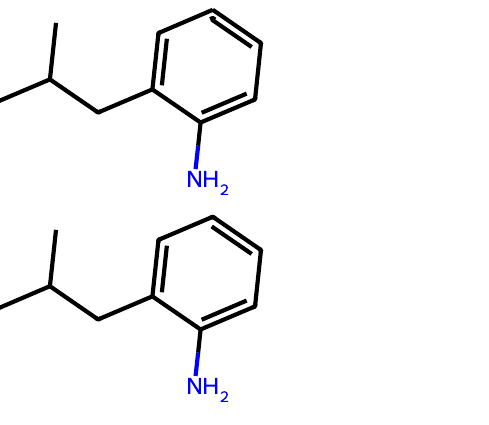}} & \rotatebox{90}{\includegraphics[width=0.15\linewidth,trim={22cm 12cm 34cm 2.9cm}, clip]{figures/case_drug.pdf}} & \includegraphics[width=0.15\linewidth,trim={30cm 13.5cm 22cm 2.9cm}, clip]{figures/case_drug.pdf} & \includegraphics[width=0.15\linewidth,trim={40cm 13.5cm 12cm 2.9cm}, clip]{figures/case_drug.pdf} & \includegraphics[width=0.15\linewidth,trim={51cm 13.5cm 0.5cm 2.9cm}, clip]{figures/case_drug.pdf}\\
        & QED: 0.79 & QED: 0.79 & QED: 0.77 & QED: 0.77 & QED: 0.63\\
        \rotatebox{90}{3M-Diffusion}&  \includegraphics[width=0.15\linewidth,trim={8cm 2cm 44cm 13cm}, clip]{figures/case_drug.pdf} &  \includegraphics[width=0.15\linewidth,trim={19cm 2cm 36cm 13cm}, clip]{figures/case_drug.pdf} &  \includegraphics[width=0.15\linewidth,trim={30cm 2cm 24cm 13cm}, clip]{figures/case_drug.pdf} &  \includegraphics[width=0.15\linewidth,trim={40cm 2cm 13cm 13cm}, clip]{figures/case_drug.pdf} &  \includegraphics[width=0.15\linewidth,trim={50cm 2cm 1cm 13cm}, clip]{figures/case_drug.pdf}\\
        & QED: 0.89 & QED: 0.83 & QED: 0.80 & QED: 0.78 & QED: 0.76\\
          
    \end{tabular}
    }
    \caption{Comparison of molecules generated by MolSnap, 3M-Diffusion, and MolT5-large under the drug likeliness condition. Top molecules are selected based on desired properties, with drug likeliness measured by QED (higher values indicate better drug likeliness).}
    \label{tab:case_drug}
\end{table*}

\section{Case Studies}
We present visual comparisons of molecules generated under different conditions. Table~\ref{tab:case_an} shows generation results under the anthocyanidin cation condition, where MolSnap consistently produces structures with higher similarity to the ground truth compared to baselines. In Table~\ref{tab:case_drug}, molecules are sampled based on the drug-likeness condition, with top samples selected by QED scores. MolSnap yields chemically diverse and property-aligned molecules with higher QED, demonstrating its superior conditional generation capability.
\end{document}